\documentclass[10pt,twocolumn,letterpaper]{article}

\usepackage[pagenumbers]{iccv} 

\usepackage{graphicx}
\usepackage{booktabs}
\usepackage{amsmath}
\usepackage{amssymb}
\usepackage{multirow}
\usepackage{array}
\usepackage{makecell}
\usepackage{bm}
\usepackage{cuted}
\usepackage{capt-of}

\definecolor{iccvblue}{rgb}{0.21,0.49,0.74}
\usepackage[pagebackref,breaklinks,colorlinks,allcolors=iccvblue]{hyperref}
\usepackage{makecell}

\def\paperID{10667} 
\def\confName{ICCV}
\def\confYear{2025}

\title{One Look is Enough: Seamless Patchwise Refinement for Zero-Shot Monocular Depth Estimation on High-Resolution Images}

\author{Byeongjun Kwon\\
KAIST\footnotemark[1]\\
{\tt\small byeongjun.kwon@kaist.ac.kr}
\and
Munchurl Kim \footnotemark[2]\\
KAIST\footnotemark[1]\\
{\tt\small mkimee@kaist.ac.kr}
\and
\small{\url{https://kaist-viclab.github.io/One-Look-is-Enough_site}}
}

\begin{document}
\maketitle
\begin{strip}\centering
    \vspace{-1.5cm}
    \includegraphics[width=1.0\linewidth]{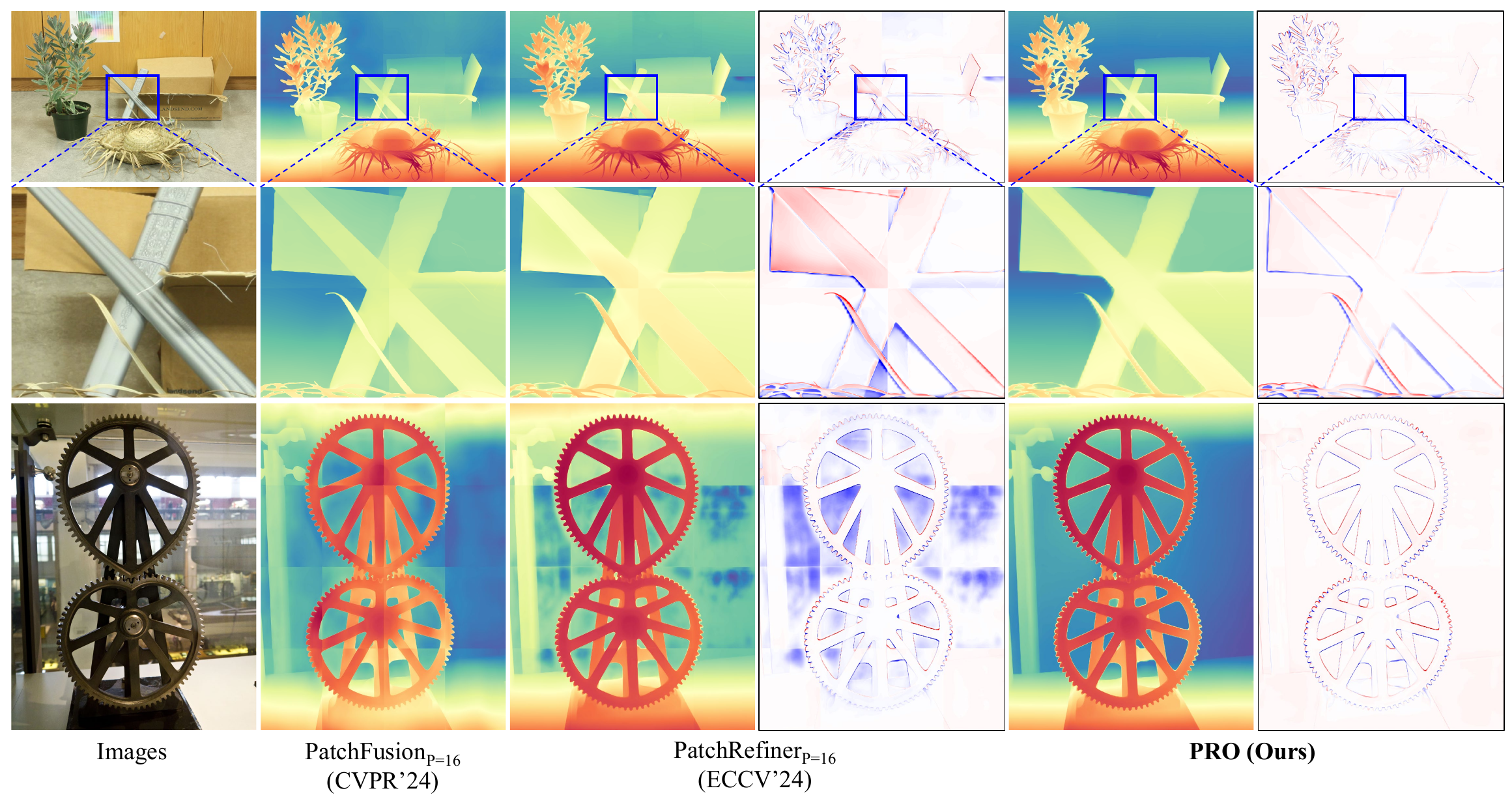}
    \vspace{-0.6cm}
    \captionof{figure}{\textbf{Qualitative comparison of patch-based depth estimation models (PatchFusion \cite{pf}, PatchRefiner \cite{pr} and PRO (Ours)) for high-resolution images.} PatchRefiner \cite{pr} and PRO (Ours) also visualize the residuals to effectively highlight depth discontinuity issues in the 4th and 6th columns. All models perform depth estimations on $4\times4$ patches of input images. Zoom-in can help distinguish the grid lines (depth discontinuities) of the $4\times4$ patches. Our proposed PRO achieves smooth boundary transitions without depth discontinuity artifacts along the grid boundaries.}
    \label{fig:first}
\end{strip}

{
  \renewcommand{\thefootnote}%
    {\fnsymbol{footnote}}
  \footnotetext[1]{Korea Advanced Institute of Science and Technology.}
  \footnotetext[2]{Corresponding author.}
}

\begin{abstract}
Zero-shot depth estimation (DE) models exhibit strong generalization performance as they are trained on large-scale datasets. However, existing models struggle with high-resolution images due to the discrepancy in image resolutions of training (with smaller resolutions) and inference (for high resolutions). Processing them at full resolution leads to decreased estimation accuracy on depth with tremendous memory consumption, while downsampling to the training resolution results in blurred edges in the estimated depth images. Prevailing high-resolution depth estimation methods adopt a patch-based approach, which introduces depth discontinuity issues when reassembling the estimated depth patches, resulting in test-time inefficiency. Additionally, to obtain fine-grained depth details, these methods rely on synthetic datasets due to the real-world sparse ground truth depth, leading to poor generalizability. To tackle these limitations, we propose Patch Refine Once (PRO), an efficient and generalizable tile-based framework. Our PRO consists of two key components: (i) Grouped Patch Consistency Training that enhances test-time efficiency while mitigating the depth discontinuity problem by jointly processing four overlapping patches and enforcing a consistency loss on their overlapping regions within a single backpropagation step, and (ii) Bias Free Masking that prevents the DE models from overfitting to dataset-specific biases, enabling better generalization to real-world datasets even after training on synthetic data. Zero-shot evaluations on Booster, ETH3D, Middlebury 2014, and NuScenes demonstrate that our PRO can be seamlessly integrated into existing depth estimation models. It preserves the performance of original depth estimation models even under grid-based inference on high-resolution images, exhibiting minimal depth discontinuities along patch boundaries. Moreover, our PRO achieves significantly faster inference speed compared to prior patch-based methods.
\end{abstract}    
\vspace{-0.3cm}
\vspace{-0.2cm}
\section{Introduction}
\label{sec:intro}

Monocular Depth Estimation (MDE) \cite{eigen2014depth,saxena2008make3d,shao2023nddepth,ground,temporal,md1,md2} has been widely investigated as the demand for 3D information continues to grow in autonomous driving, robotics, and virtual reality applications. After Midas \cite{midas} proposed an MDE network trainable with mixed training datasets, numerous zero-shot MDE networks \cite{dpt, omni, midav3,da,da2, marigold, geowizard} have been studied to predict depth for unseen real-world images. However, due to the architectural limitation or limited image resolutions of training datasets, most zero-shot MDE networks are trained specifically with the images of low resolutions (e.g. 384$\times$384, 518$\times$518). Consequently, when high resolution images are fed into these models, the resulting estimated depth images tend to contain disrupted overall structures with low-frequency artifacts, although yielding improved edge details \cite{boost,gbdf}.

Patch-based methods \cite{boost,pr,pf} have shown promising results by splitting high-resolution images into patches for DE, mitigating memory consumption issues while achieving remarkable performance. However, they suffer from depth discontinuity problems (e.g. boundary artifacts along grid boundaries) which occur when independent DE patches are reassembled to construct a complete (whole) depth map because depth continuity is maintained within each patch but not between patches. Previous methods \cite{pr,pf} alleviate this depth discontinuity problem (i) by incorporating a consistency loss during training and (ii) by ensemble averaging at test time which slows down inference speed, making it impractical for real-world applications.

\begin{figure}[tbp]
  \centering
  \includegraphics[width=1.0\columnwidth]{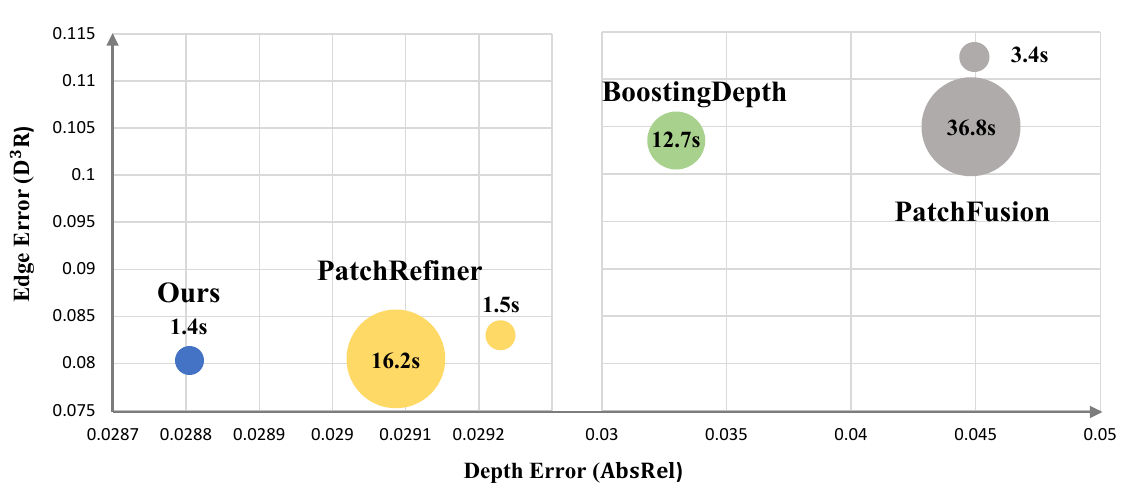}
  \vspace{-0.7cm}
  \caption{\textbf{Comparison of performance and efficiency on Middlebury 2014 \cite{middle}.} The area of each circle represents the inference time. Circles of the same color represent the same model with different patch numbers ($\mathrm{P}=16$ for small-sized yellow and grey circles, $\mathrm{P}=177$ for large-sized yellow and grey circles). Our model (PRO) achieves the best performance in terms of edge errors ($\mathrm{D}^3\mathrm{R}$) and depth errors (AbsRel) while maintaining the fastest inference time. Inference time for all models is measured on the RTX 4090.}
  \label{fig:time}
  \vspace{-0.5cm}
\end{figure}

A common strategy for training high-resolution depth estimation is to train models on real-world high-resolution datasets \cite{holopix,ibims, wsvd,vdw, middle} or leverage high-resolution synthetic datasets \cite{smd,tartanair,irs,mvssynth}. Real-world datasets have the advantage of a smaller domain gap compared to synthetic datasets, but their ground truth (GT) depths are often sparse, particularly around edges, where supervision is crucial. Additionally, the difficulty of acquiring dense depth data in real-world scenarios hinders the training of DE models for high-resolution images. On the other hand, synthetic datasets provide dense GT depth with fine details, but models trained on synthetic datasets often struggle with zero-shot inference on real-world data due to the domain gap \cite{sddr}. Moreover, we observe that transparent objects are inaccurately labeled in the UnrealStereo4K dataset \cite{smd}, the only synthetic dataset providing 4K depth annotations, as their depth values correspond to the background behind them.

To address the aforementioned problems, we propose Patch Refine Once (PRO), a novel refinement model that enables DE models to generate accurate depth predictions with only a single refinement per patch at inference time (i.e.,``One Look''). Our PRO consists of two strategies: Grouped Patch Consistency Training and Bias Free Masking. Fig.~\ref{fig:first} shows a qualitative comparison of patch-based depth estimation models for high-resolution images. As shown, our PRO yields depth refinement results with minimal depth discontinuities along grid boundaries, whereas two recent state-of-the-art (SOTA) models still exhibit noticeable boundary artifacts. Fig.~\ref{fig:time} illustrates a comparison of performance and efficiency between recent SOTA depth refinement models and our PRO. As observed, our PRO provides the best performance in terms of $\mathrm{D}^3\mathrm{R}$, AbsRel and inference speed. Our contribution is summarized as:

\begin{itemize}
    \item We propose Grouped Patch Consistency Training (GPCT) strategy that can be well harmonized with existing DE models to mitigate the depth discontinuity problem for patch-wise DE approaches on high-resolution images. The GPCT does not require test-time ensemble, allowing a 12$\times$ faster inference speed compared to the SOTA patch-based method. 
    \item We introduce Bias Free Masking (BFM) to selectively apply supervision signals by masking out unreliable regions in images. The unreliable regions are the regions that have inaccurately annotated depth in GT, which is often observed in the UnrealStereo4K dataset. This issue is effectively addressed by our BFM, which prevents the model from overfitting to domain-specific biases.
    \item Our PRO achieves SOTA performance in zero-shot evaluation on Booster \cite{booster}, ETH3D \cite{eth}, Middlebury 2014 \cite{middle}, and NuScenes \cite{nuscenes} across all metrics, and achieves efficient inference (12$\times$ faster than PatchRefiner with 177 patches), outperforming recent state-of-the-art depth refinement methods on high-resolution images.

\end{itemize}

\section{Related work}
\label{sec:related}
\subsection{Zero-shot Monocular Depth Estimation}
Early MDE models \cite{adabins,binsformer,eigen2014depth, saxena2008make3d,newcrf} were trained individually on each dataset \cite{kitti,nyuv2}, achieving high performance on the training data but poor generalization to unseen datasets due to the domain gap. Zero-shot MDE models that perform well on unseen images have been widely studied to improve generalization. MegaDepth \cite{mega} and DiverseDepth \cite{diverse} construct large-scale datasets by collecting Internet images, improving adaptability to images from diverse scenes and conditions. They have inspired further efforts to scale up training datasets to enhance zero-shot performance. MiDaS \cite{midas} proposes a scale and shift invariant loss, which enables MDE networks to be trained on a diverse mixture of datasets, making the model robust to unseen datasets. DPT \cite{dpt}, Omnidata \cite{omni}, and MiDaS v3.1 \cite{midav3} improve DE performance by replacing CNNs with transformer architectures. DepthAnything \cite{da} introduces a semi-supervised approach that utilizes 62M unlabeled images as pseudo labels for training. This substantial increase in data scale has led to improved generalization performance. DepthAnythingV2 \cite{da2} observes that real-world depth datasets contain label noise and lack fine details. It is trained with pseudo GT labels obtained from a teacher model trained on a synthetic dataset, achieving high-quality depth maps. 

Instead of scaling up datasets, some studies utilize prior knowledge from diffusion models for DE. Marigold \cite{marigold} fine-tunes a pretrained Stable Diffusion \cite{ldm} model on a relatively small synthetic dataset, showing competitive results. Geowizard \cite{geowizard} jointly estimates depth and surface normals by leveraging self-attention across different domains to enhance geometric consistency.
However, as most zero-shot MDE models are trained on low-resolution images (e.g., 384$\times$384, 518$\times$518), their performance degrades when applied to high-resolution images. When high-resolution images are downsampled to the training resolution for inference, fine details are lost. On the other hand, performing inference with a larger resolution input preserves details but degrades depth accuracy.

\subsection{High-Resolution Depth Estimation}
High-resolution DE models aim to predict accurate depth while preserving fine details. SMD-Net \cite{smd} utilizes an implicit function \cite{inr} to predict a mixture density for precise DE at object boundaries. Dai \emph{et al.} \cite{gbdf} introduce a Poisson fusion-based depth map optimization method by applying guided filtering in a self-supervised learning framework. SDDR \cite{sddr} proposes a self-distilled depth refinement method that generates pseudo edge labels, addressing local consistency issues and edge deformation noise through a noisy Poisson fusion process. 
Patch-based high-resolution DE methods select patches and merge patch-wise results to improve depth details. BoostingDepth \cite{boost} introduces a patch selection process based on edge density, and employs iterative refinement through multi-resolution depth merging. PatchFusion \cite{pf} avoids patch selection by applying a predefined partitioning scheme, and leverages shifted window self-attention to fuse global and local information. PatchRefiner \cite{pr} initializes depth estimation with a low-resolution depth map, predicts residuals, and proposes a detail and disentangling (DSD) loss for training on both real-world and synthetic datasets. However, patch-based methods inherently suffer from depth discontinuities at patch boundaries, as they emphasize local information over global consistency, or maintain depth continuity only within individual patches. To address this problem, prior works \cite{pf,pr} employ a large number of patches for test-time ensemble, which significantly slows down inference. By contrast, our PRO (Patch Refine Once) model efficiently achieves fine-grained depth refinement with only a single pass per patch during inference, despite employing a tile-based approach.

\subsection{Training Data for HR Depth Estimation}
High-resolution DE requires datasets with fine-grained depth details and minimal label noise. Real-world datasets \cite{kitti,nyuv2, nuscenes, guizilini20203d, mega} are often sparse due to the limitations of LiDAR or contain inaccurate depth boundaries when obtained through Structure-from-Motion (SfM). Therefore, previous research \cite{gbdf,boost} has trained depth networks using a small number of relatively dense real-world datasets \cite{hr-wsi,ibims,middle}. Some studies \cite{gbdf, sddr} utilize pseudo labels for self-supervision, but these labels often contain noise. Other works \cite{pf, pr} leverage synthetic datasets for more accurate depth, but suffer from poor generalization ability due to the domain gap. To take advantage of dense depth annotations in synthetic datasets without compromising generalization performance, we introduce Bias Free Masking (BFM), which selectively identifies the regions where supervision is applied. It mitigates overfitting to biases in synthetic datasets and preserves robustness in zero-shot high-resolution DE.

\begin{figure}[t]
  \centering
  \includegraphics[width=0.9\columnwidth]{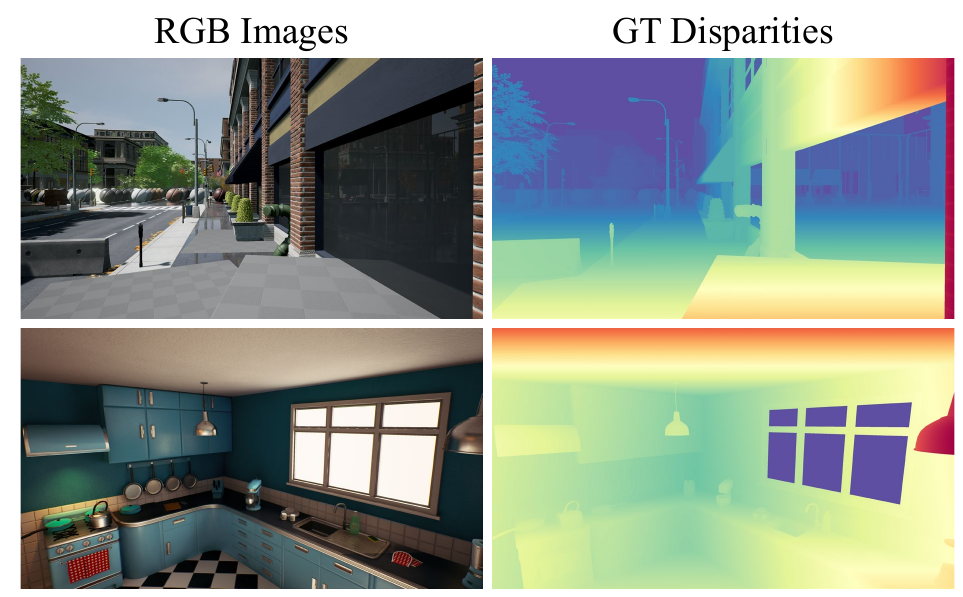}
  \vspace*{-3mm}
  \caption{Labeling error examples in UnrealStereo4K \cite{smd} dataset.}
  \label{fig:ex}
\end{figure}
\vspace*{-1.1cm}

\section{Method}
\label{sec:method}


\begin{figure*}[tbp]
  \centering
  \includegraphics[width=1.0\textwidth]{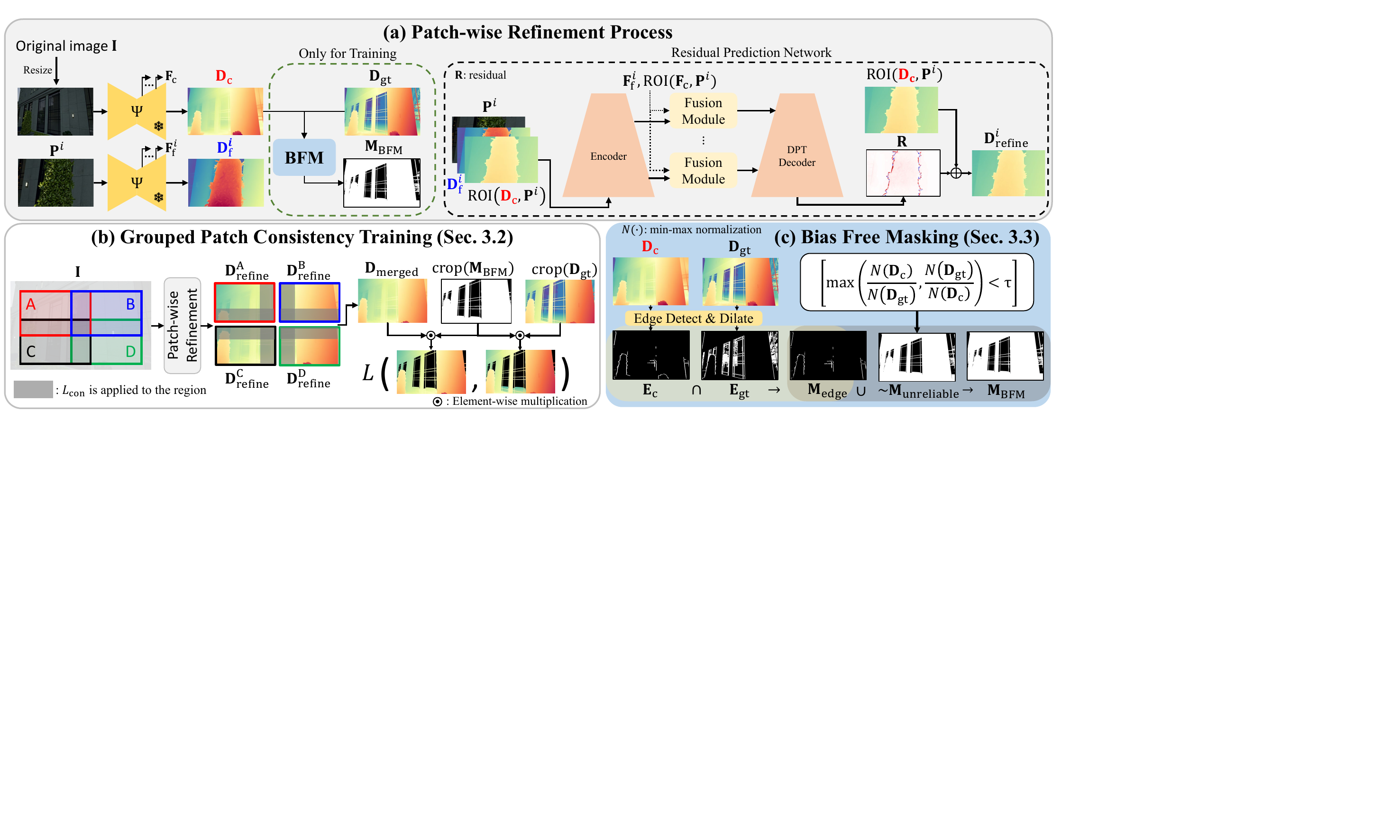}
  \vspace{-0.6cm}
  \caption{\textbf{Overview of the Framework.} \textbf{(a) Patch-wise Refinement Process.} First, the pretrained MDE model $\Psi$ estimates a coarse depth $\mathbf{D}_\mathrm{c}$ and a fine depth $\mathbf{D}_\mathrm{f}^{i}$ from the resized input image $\mathbf{I}$ and the $i$-th patch $
  \mathbf{P}^i$. Based on $\mathbf{D}_\mathrm{c}$, the residual prediction network predicts residuals $\mathbf{R}$. This process is applied to every patch. Note that Bias Free Masking (BFM) is only required at training. \textbf{(b) Grouped Patch Consistency Training.} The training image is cropped with overlapping regions, followed by patch-wise refinement. Subsequently, depth consistency loss $\mathcal{L}_{\mathrm{con}}$ is applied to the gray-shaded regions to enforce consistency between depth results refined separately. \textbf{(c) Bias Free Masking.} We identify the reliable region as $\mathbf{M}_{\mathrm{BFM}}$ by utilizing $\mathbf{D}_\mathrm{c}$ and $\mathbf{D}_\mathrm{gt}$.}
  \label{fig:main}
  \vspace{-0.2cm}
\end{figure*}

We describe our Patch Refine Once (PRO) model. Section \ref{3.1} introduces the overall pipeline of PRO, followed by the two key contributions: Grouped Patch Consistency Training (GPCT) and Bias Free Masking (BFM), which are detailed in Sections \ref{gpct} and \ref{mask}, respectively.

\subsection{Overview of Patch Refine Once (PRO)}\label{3.1}
Fig.~\ref{fig:main} illustrates the overall pipeline of the proposed PRO framework. Following \cite{pf}, given an original input image $\mathbf{I} \in \mathbb{R}^{H\times W \times 3}$, we estimate a coarse depth map $\mathbf{D}_\mathrm{c}$ by feeding a resized version of $\mathbf{I}$ into a pretrained zero-shot MDE network $\Psi$. $\mathbf{D}_\mathrm{c}$ captures the overall scene structure, which is essential for preserving global consistency. 

To complement the local details, we take the following steps. First, we crop the original image $\mathbf{I}$ into $N$ grid-based patches $\{\mathbf{P}^{i}\}_{i=1}^{N}$. Each individual patch $\mathbf{P}^{i}$ is then resized and fed into $\Psi$ to estimate the corresponding fine depth map $\mathbf{D}_{\mathrm{f}}^{i}$. We extract two sets of five-level features from the decoder of $\Psi$: (i) $\mathbf{F}_\mathrm{c} = \{\mathbf{f}_{\mathrm{c}, j}^{i}\}_{j=1}^{5}$ for the resized image, and (ii) $\mathbf{F}_\mathrm{f} = \{\mathbf{f}_{\mathrm{f},j}^{i}\}_{j=1}^{5}$ for the patch $\mathbf{P}^{i}$. To align $\mathbf{F}_\mathrm{c}$ with $\mathbf{F}_\mathrm{f}$, we apply the $\mathsf{ROI}$ operation \cite{maskrcnn} to extract patch-aligned features from $\mathbf{F}_\mathrm{c}$. The patch-level features $\mathbf{F}_{\mathrm{f}}^{i}$ and the corresponding $\mathsf{ROI}$-extracted features are subsequently passed into a fusion module within the residual prediction network $\theta$. The fusion module integrates these features using a wavelet transform \cite{daubechies1990wavelet} and shallow convolutional layers, effectively combining global and local information. See \emph{supplementary material} for details of the fusion module. Finally, the refined depth map for $\mathbf{P}^{i}$, denoted as $\mathbf{D}_{\mathrm{refine}}^{i}$, is obtained as:
\begin{equation}
\mathbf{D}_{\mathrm{refine}}^{i} = \theta(\mathsf{concat}(\mathbf{P}^{i}, \mathsf{ROI}(\mathbf{D}_{\mathrm{c}},\mathbf{P}^{i}), \mathbf{D}_{\mathrm{f}}^{i})),
\end{equation}
where $\mathsf{ROI}(\mathbf{D}_{\mathrm{c}}, \mathbf{P}^{i})$ extracts the region of interest from $\mathbf{D}_{\mathrm{c}}$ corresponding to the spatial extent of $\mathbf{P}^{i}$, yielding $\mathbf{D}_{\mathrm{c}}^{i}$. The operator $\mathsf{concat}$ denotes channel-wise concatenation. Unlike previous methods \cite{pf,pr} where separate networks are trained to predict $\mathbf{D}_\mathrm{c}$ and $\mathbf{D}_\mathrm{f}$ individually, we utilize the same pretrained zero-shot MDE network $\Psi$ and train only the residual prediction network as shown in Fig.~\ref{fig:main}-(a).

\subsection{Grouped Patch Consistency Training (GPCT)}\label{gpct}
To address the boundary artifacts introduced by patch-wise refinement, we propose a simple yet effective GPCT strategy that ensures depth consistency across patch boundaries. While previous approaches \cite{pf,pr} focus only on two diagonally adjacent overlapping patches (e.g., (A, D) or (B, C) in Fig.~\ref{fig:main}-(b)) to enforce consistency constraints, our method employs four overlapping patches (A, B, C, and D in Fig.~\ref{fig:main}-(b)) simultaneously. As shown in Fig.~\ref{fig:main}-(b), we divide the training sample into overlapping patches, so that each patch overlaps with its neighbors. After refining each patch independently, we apply a depth consistency loss $\mathcal{L}_{\text{con}}$ to enforce consistency between the depth refinement results of overlapping patches:
\vspace{-0.2mm}
\begin{equation}
\mathcal{L}_{\text{con}} = \sum_{i \neq j} \frac{1}{|\Omega|}  \sum_{p \in \Omega}  \left( \mathbf{D}_{\mathrm{refine}}^i(p) - \mathbf{D}_{\mathrm{refine}}^j(p) \right)^2,
\end{equation}
where $ \mathbf{D}_{\mathrm{refine}}^i $ and $ \mathbf{D}_{\mathrm{refine}}^j $ denote the depth predictions from overlapping patches $i$ and $j$, respectively. We denote the overlapping region between patches $i$ and $j$ as $\Omega$, with $|\Omega|$ indicating the number of pixels within this region.
The merged depth map $ \mathbf{D}_{\text{merged}} $ at position $p$ is computed as:
\begin{equation}
\mathbf{D}_{\mathrm{merged}}(p) = \frac{1}{N_{\mathrm{o}}(p)}\sum_{i \in \{A, B, C, D\}} \mathbf{D}_{\mathrm{refine}}^i(p),
\end{equation}
where $\mathbf{D}_{\mathrm{refine}}^i$ denotes the refined depth prediction from patch $i$ (i.e., $i \in \{A, B, C, D\}$ as shown in Fig.~\ref{fig:main}-(b)), and $N_{\text{o}}(p)$ is the number of overlapping patches that contribute a depth estimate at pixel $p$.
Then, we calculate the loss between merged depth $\mathbf{D}_{\mathrm{merged}}$ and GT depth $\mathbf{D}_{\mathrm{gt}}$.

Since our method computes the loss using all four patches simultaneously, every backward propagation step enforces consistency at all boundaries. In contrast, the prior method \cite{pf} only applied a consistency loss on the overlapped region between two patches (e.g., (A, D) or (B, C)) at a time, leading to insufficient boundary supervision. Our approach provides a stronger supervision signal, yielding significantly improved cross-patch consistency and smooth boundary transitions during inference. By training with this GPCT strategy, our PRO model mitigates boundary artifacts without requiring additional refinement during inference, allowing efficient patch-wise depth estimation, as only a single refinement step per patch (i.e., ``One Look'') is required.

\subsection{Bias Free Masking (BFM)}\label{mask}
To utilize dense GT depth when training depth refinement models for fine-grained details, we use a synthetic dataset such as the UnrealStereo4K \cite{smd}. However, the UnrealStereo4K dataset, which provides 4K resolutions, annotates the depths of transparent objects as the depths of their backgrounds (e.g., window objects in Fig.~\ref{fig:ex}), which introduces a specific bias in the dataset. To maintain the benefits of dense depth labeling while avoiding overfitting to the dataset-specific biases, we propose Bias Free Masking (BFM) that uses the prior knowledge of the pretrained zero-shot MDE models. That is, BFM aims to exclude the unreliable regions corresponding to the incorrect GT depths in transparent objects in the synthetic dataset (UnrealStereo4K) during training. Since the pretrained zero-shot MDE model $\Psi$ can be considered to have informative prior knowledge learned from large scale image-depth pairs, the model can be used to identify unreliable regions where $ \mathbf{D}_{\mathrm{c}} $ and $ \mathbf{D}_{\mathrm{gt}} $ significantly deviate from each other, indicating potential biases to the synthetic dataset.

\begin{table*}[tbp]
    \scriptsize
    \centering
    \resizebox{1.0\textwidth}{!}{
    \def\arraystretch{1.2}
    \begin{tabular} {c|c|c|cc|cc|ccc|cc|c}
        \Xhline{2\arrayrulewidth}
        \multirow{2}{*}{Methods} & \multirow{2}{*}{Publications} & \multirow{2}{*}{Runtime (s)}  & \multicolumn{2}{c|}{\textbf{Booster}} & \multicolumn{2}{c|}{\textbf{ETH3D}} & \multicolumn{3}{c|}{\textbf{Middle14}} & \multicolumn{2}{c|}{\textbf{NuScenes}} & \textbf{DIS}\\
        \cline{4-13}
        & & & AbsRel$\downarrow$ & $\delta_1\uparrow$ & AbsRel$\downarrow$ & $\delta_1\uparrow$ & AbsRel$\downarrow$ & $\delta_1\uparrow$ & $\mathrm{D}^3\mathrm{R}\downarrow$ & AbsRel$\downarrow$ & $\delta_1\uparrow$ & BR$\uparrow$ \\
        \hline
        DepthAnythingV2 \cite{da2} & NIPS 2024 & - & 0.0307 & 0.993 & 0.0465 & 0.983 & 0.0307 & 0.994 & 0.0979 &0.106 & 0.882 & 0.059 \\
        BoostingDepth \cite{boost}   & CVPR 2021 & 12.7 & 0.0330 & 0.993 & 0.0552 & 0.974 & 0.0330 & 0.995 & 0.1035 & 0.115 & 0.870  & 0.170\\
        PatchFusion $_{\mathrm{P}=16}$ \cite{pf}   & CVPR 2024 & 3.4 & 0.0504 & 0.985 & 0.0735 & 0.956 & 0.0450 & 0.989 & 0.1124 & 0.141 & 0.831 & \textbf{0.206} \\
        PatchFusion $_{\mathrm{P}=177}$ \cite{pf}   & CVPR 2024 & 36.8 & 0.0496 & 0.986 & 0.0723 & 0.957 & 0.0448 & 0.989 & 0.1050 & 0.139 & 0.833 & 0.189\\
        PatchRefiner$_{\mathrm{P}=16}$ \cite{pr}   & ECCV 2024 & 1.5 &  0.0348 & 0.989 & 0.0435 & \textbf{0.985} & 0.0292 & 0.995 & 0.0830  & 0.107 & 0.879 & 0.151\\
        PatchRefiner$_{\mathrm{P}=177}$ \cite{pr}   & ECCV 2024 & 16.2 & 0.0336 & 0.991 & 0.0430 & \textbf{0.985} & 0.0292 & 0.995 & 0.0805  & 0.106 & 0.881 & 0.141\\
        \textbf{PRO (Ours)} & - & {\textbf{1.4}} & {\textbf{0.0304}} & {\textbf{0.994}} & {\textbf{0.0422}} & {\textbf{0.985}} & {\textbf{0.0287}} & {\textbf{0.996}} & {\textbf{0.0803}} & {\textbf{0.104}} & {\textbf{0.883}} & 0.156\\
        \Xhline{2\arrayrulewidth}
    \end{tabular}}
    \vspace{-0.2cm}
    \caption{Quantitative comparison of depth estimation methods on Booster \cite{booster}, ETH3D \cite{eth}, Middlebury 2014 \cite{middle}, NuScenes \cite{nuscenes}, and DIS-5K \cite{dis} datasets. \textbf{Bold} indicates the best performance in each metric.}
    \vspace{-0.3cm}
    \label{tab:main}
\end{table*}

Given a resized version for input image $\mathbf{I}$, we obtain coarse depth $ \mathbf{D}_{\mathrm{c}} $ from $\Psi$ and GT depth $ \mathbf{D}_{\mathrm{gt}} $ from the synthetic dataset. Then, by measuring the relative consistency between $ \mathbf{D}_{\mathrm{c}} $ and $ \mathbf{D}_{\mathrm{gt}} $, we identify unreliable regions as:
\begin{equation}
\mathbf{M}_{\mathrm{unreliable}} =  \left[ \max \left( \frac{N(\mathbf{D}_{\mathrm{c}})}{N(\mathbf{D}_{\mathrm{gt}})}, \frac{N(\mathbf{D}_{\mathrm{gt}})}{N(\mathbf{D}_{\mathrm{c}})} \right) > \tau \right],
\end{equation}
where $\left[ \cdot \right]$ represents the Iverson bracket, $N(\mathbf{D})$ denotes min-max normalization for $\mathbf{D}$  and $ \tau $ is an empirically chosen threshold (set to 2 in our experiments). $\mathbf{M}_{\mathrm{unreliable}}$ as an unreliable mask identifies the regions where $ \mathbf{D}_{\mathrm{c}} $ and $ \mathbf{D}_{\mathrm{gt}} $ significantly deviate, indicating potential synthetic dataset biases or inconsistencies, as aforementioned. Since $ \mathbf{D}_{\mathrm{c}} $ lacks sharp edges, complementing $\mathbf{M}_{\mathrm{unreliable}} $ to obtain the reliable regions is a naive approach, which causes the exclusion of critical edge regions from training, thus resulting in the removal of valuable depth gradients required for refinement. To address this, we incorporate edge information into the reliable mask generation. However, simply utilizing the edge map from $\mathbf{D}_{\mathrm{gt}}$ as an edge mask could bring noise in a training sample because the flat transparent region could include the edges due to background with edges as observed in Fig.~\ref{fig:main}-(c). To handle this problem, we additionally employ the edge maps from $\mathbf{D}_{\mathrm{c}}$. Consequently, we first extract edges from both $ \mathbf{D}_{\mathrm{c}} $ and $ \mathbf{D}_{\mathrm{gt}} $, and dilate them as:
\begin{equation}
\mathbf{E}_{\mathrm{c}} = \text{dilate}(\text{edge}(\mathbf{D}_{\mathrm{c}})), \quad \mathbf{E}_{\mathrm{gt}} = \text{dilate}(\text{edge}(\mathbf{D}_{\mathrm{gt}})),
\end{equation}
to guarantee that the final edge mask should not miss the regions that require depth refinements, otherwise they might be excluded simply from complementing $\mathbf{M}_{\mathrm{unreliable}} $.
We define the edge mask as $\mathbf{M}_{\mathrm{edge}} = \mathbf{E}_{\mathrm{c}} \cap \mathbf{E}_{\mathrm{gt}}$.
Finally, the reliable mask for BFM is constructed as $\mathbf{M}_{\mathrm{BFM}} = \textbf{M}_{\mathrm{edge}} \cup \sim \textbf{M}_{\mathrm{unreliable}}$ where $\sim \textbf{M}_{\mathrm{unreliable}}$ is the complement of $\textbf{M}_{\mathrm{unreliable}}$. This ensures that the unreliable regions caused by large discrepancies between $ \mathbf{D}_{\mathrm{c}} $ and $ \mathbf{D}_{\mathrm{gt}} $ are ignored, while essential edge details are preserved for training. Consequently, supervision is applied only to the reliable regions via a masked loss:
\begin{equation} \label{mask_loss}
\mathcal{L}_{\mathrm{masked}} = \frac{1}{|\textbf{M}_{\mathrm{BFM}}|} \sum_{p \in \textbf{M}_{\mathrm{BFM}}} \mathcal{L}(\mathbf{D}_{\mathrm{merged}}(p), \mathbf{D}_{\mathrm{gt}}(p))
\end{equation} 
where $ \mathcal{L} $ is the loss between merged depth and GT depth, which is the combination of an L1 loss, an L2 loss, and a multi-scale gradient loss with four scale levels \cite{mega} using a ratio of 1:1:5.
By restricting training to reliable regions, the depth refinement model refines depth only where necessary, avoiding modifications in regions where the pretrained model's prior knowledge is considered more reliable than the (noisy) GT. This helps prevent the model from refining transparent regions incorrectly.

\noindent\textbf{Final Loss function.} The final training loss consists of masked loss $\mathcal{L}_{\mathrm{masked}}$ and depth consistency loss $\mathcal{L}_{\mathrm{con}}$ as:
\begin{equation}
\mathcal{L}_{\mathrm{final}} = \mathcal{L}_{\mathrm{masked}} + \lambda \mathcal{L}_{\mathrm{con}}.
\end{equation}\label{final_loss}
where $\lambda$ is empirically set to 4 in our experiments.
\section{Experiments}
\label{sec:experiment}

\subsection{Datasets and Metrics}
We train recent SOTA depth refinement models and our PRO model on a synthetic dataset (UnrealStereo4K) to get the advantage of dense GT depth for a fair comparison. At test time, we evaluate its zero-shot performance on four real datasets to demonstrate generalizability.

\noindent \textbf{UnrealStereo4K.} The UnrealStereo4K dataset \cite{smd} is a synthetic dataset comprising stereo images with 4K resolution (2,160$\times$3,840) and pixel-wise GT disparity labels. As aforementioned, its some GT labels are inaccurate especially for transparent objects, as shown in Fig.~\ref{fig:ex}.

\noindent \textbf{Booster.} The Booster dataset \cite{booster} has high-resolution (3,008 $\times$ 4,112) indoor images with specular and transparent surfaces. We use the training set with GT depth for testing. To ensure diversity, we remove images captured from the same scenes, resulting in a test set of 38 images.

\noindent \textbf{ETH3D.} The ETH3D dataset \cite{eth} contains high-resolution indoor and outdoor images (6,048 $\times$ 4,032) with GT depth captured by LiDAR sensors. In the dataset, 34 images were rotated 90 degrees counterclockwise, causing the bottom of the scenes to appear on the right. Before testing, we rotate these images 90 degrees clockwise to ensure that the floor sides are correctly positioned at the bottom sides.

\noindent \textbf{Middlebury 2014.} The Middlebury 2014 dataset \cite{middle} consists of 23 high-resolution (nearly 4K) indoor images with dense ground truth disparity maps.

\noindent \textbf{NuScenes.} The NuScenes dataset \cite{nuscenes} is an autonomous driving dataset that provides multi-directional camera images along with depth information obtained from LiDAR and radar sensors. Among outdoor datasets, the Cityscapes dataset \cite{Cityscapes} provides higher resolution but suffers from blurred edges since its disparity maps are generated using a stereo matching algorithm instead of LiDAR sensors. Therefore, we use the NuScenes dataset for testing.

\noindent \textbf{DIS-5K.} The DIS-5K dataset \cite{dis} is a high-resolution image segmentation dataset that provides accurately annotated masks, making it suitable for evaluating edge accuracy.

\noindent \textbf{Evaluation Metrics.} We use the commonly adopted metrics for DE, Absolute Relative Error (AbsRel) and $\delta_1$. Additionally, to evaluate the edge quality, we adopt $\mathrm{D}^3\mathrm{R}$ metric on the Middlebury 2014 dataset following \cite{boost}, and measure Boundary Recall (BR) metric on the DIS-5K dataset following \cite{pro}. We utilize Consistency Error (CE) \cite{pf} to evaluate the consistency of the DE results across patches.

\begin{figure*}
    \centerline{\includegraphics[width=0.90\textwidth]{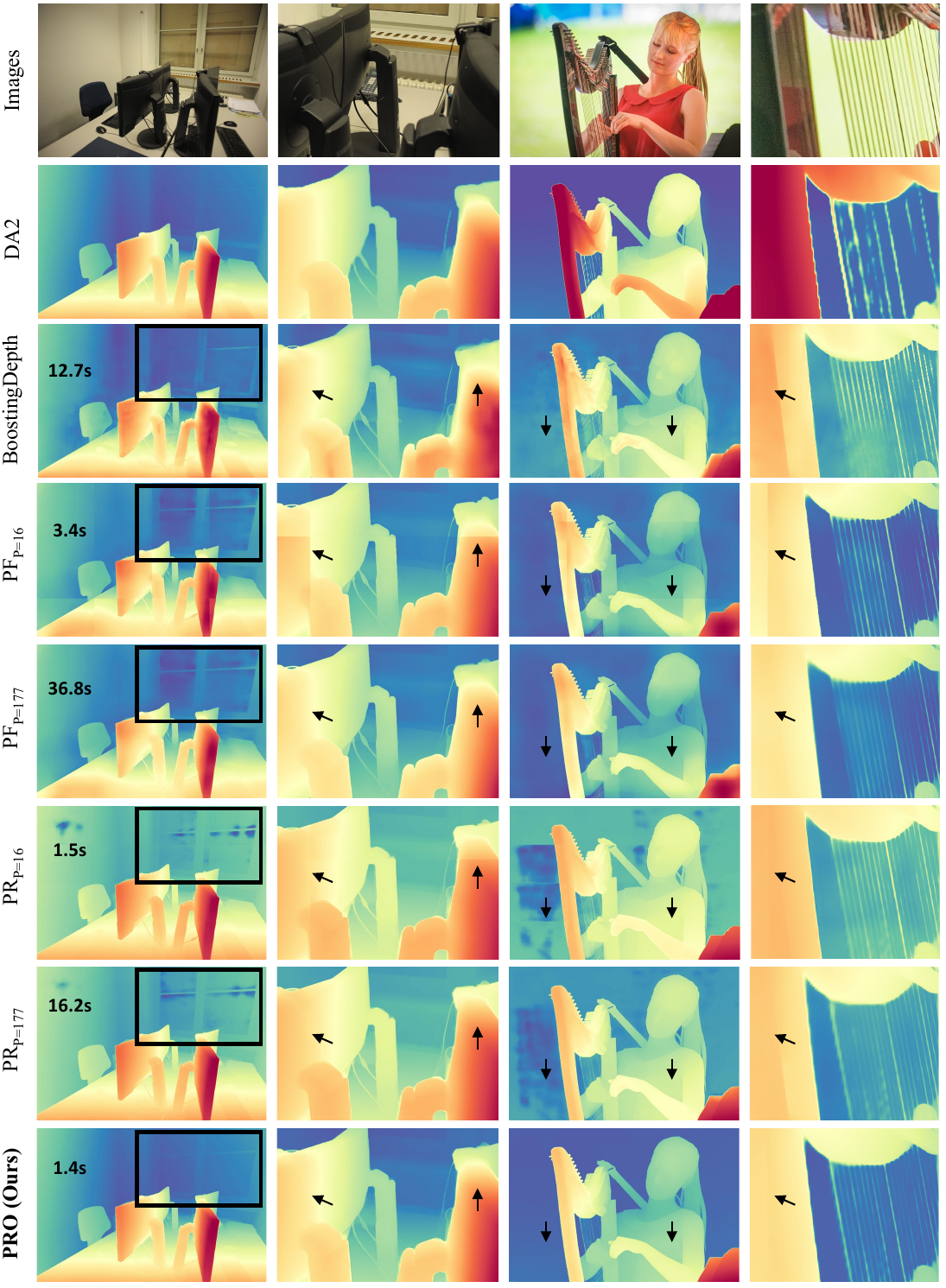}}
    \caption{\textbf{Qualitative comparisons for patch-wise DE methods on ETH3D \cite{eth} and DIS-5K \cite{dis}.} We compare PRO (Ours) with BoostingDepth \cite{boost}, PatchFusion (PF) \cite{pf}, and PatchRefiner (PR) \cite{pr}. The time displayed in the leftmost depth column represents the inference time. Black rectangle area represents the transparent object region. Black arrows indicate patch boundaries. Zoom in for details.} 
    \label{fig:qual}
\end{figure*}

\begin{table}[t]
    \scriptsize
    \centering
    \def\arraystretch{1.2}
    \begin{tabular} {c|c|c|c}
        \Xhline{2\arrayrulewidth}
        Metric & PatchFusion \cite{pf} & PatchRefiner \cite{pr} & PRO (Ours) \\
        \hline
        CE$\downarrow$ & 0.364 & 0.347 & \textbf{0.049} \\
        \Xhline{2\arrayrulewidth}
    \end{tabular}
    \vspace{-0.2cm}
    \caption{Comparison of PatchFusion \cite{pf}, PatchRefiner \cite{pr}, and PRO (Ours) in terms of Consistency Error (CE).}
    \vspace{-0.3cm}
    \label{tab:consistency}
\end{table}

\begin{table*}[tbp]
    \scriptsize
    \centering
    \resizebox{0.85\textwidth}{!}{
    \def\arraystretch{1.2}
    \begin{tabular} {c|c|c|cc|cc|ccc|cc|c}
        \Xhline{2\arrayrulewidth} \multirow{2}{*}{Model}
        & \multirow{2}{*}{GPCT} & \multirow{2}{*}{BFM} & \multicolumn{2}{c|}{\textbf{Booster}} & \multicolumn{2}{c|}{\textbf{ETH3D}} & \multicolumn{3}{c|}{\textbf{Middle14}} & \multicolumn{2}{c|}{\textbf{NuScenes}} & \multirow{2}{*}{CE$\downarrow$} \\
        \cline{4-12}
        & & & AbsRel$\downarrow$ & $\delta_1\uparrow$ & AbsRel$\downarrow$ & $\delta_1\uparrow$ & AbsRel$\downarrow$ & $\delta_1\uparrow$ & $\mathrm{D}^3\mathrm{R} \downarrow$ & AbsRel$\downarrow$ & $\delta_1\uparrow$ \\
        \hline
        (a) & &  & 0.0385 & 0.987 & 0.0428 & \textbf{0.985} & 0.0292 & 0.995 & 0.0807 & 0.105 & 0.882 & 0.208 \\
        (b) & & \checkmark & \textbf{0.0303} & \textbf{0.994} & 0.0426 & \textbf{0.985} & 0.0290 & 0.995 & 0.0837 & 0.105 & 0.882 & 0.117 \\
        (c) & \checkmark &  & 0.0313 & 0.993 & 0.0425 & \textbf{0.985} & 0.0288 & \textbf{0.996} & \textbf{0.0792} & 0.105 & 0.882 & 0.058 \\
        (d) & \checkmark & \checkmark & 0.0304 & \textbf{0.994} & \textbf{0.0422} & \textbf{0.985} & \textbf{0.0287} & \textbf{0.996} & 0.0803 & \textbf{0.104} & \textbf{0.883} & \textbf{0.049} \\
        \Xhline{2\arrayrulewidth}
    \end{tabular}}
    \vspace{-0.2cm}
    \caption{Ablation on GPCT and BFM of our PRO for Booster \cite{booster}, ETH3D \cite{eth}, Middlebury 2014 \cite{middle}, and NuScenes \cite{nuscenes} datasets.}
    \vspace{-0.2cm}
    \label{tab:ablation}
\end{table*}

\subsection{Implementation Details}
We adopt the pretrained DepthAnythingV2 (DA2) \cite{da2} as the baseline model ($\Psi$ in Fig.~\ref{fig:main}). PRO, PatchFusion \cite{pf}, and PatchRefiner \cite{pr} are (re)trained based on DA2, while BoostingDepth \cite{boost} uses DA2 without additional training. The input resolution to $\Psi$ is fixed at $518\times518$. When we use  the GPCT strategy, adjacent patches are cropped with an overlap of 224 pixels, ensuring an overlap of  $h\times 224$ for horizontally adjacent patches and $224\times w$  for vertically adjacent patches, where $h$ and $w$ denote the height and width of the patch, respectively. In our BFM, the dilation kernel size is empirically set to (10, 20). Training is conducted on 7,592 UnrealStereo4K samples for 8 epochs with a batch size of 64 (with gradient accumulation), taking 10 hours on a single RTX 4090 GPU. Additional details are described in \emph{supplementary material}.

\subsection{Performance Comparison} \label{perf_comp}
At test time, each input image is divided into a $4 \times 4$ grid of patches. The depth of each patch is refined individually, and the refined patches are then reassembled to generate the final depth map.


\noindent\textbf{Zero-shot performance on high-resolution datasets.} We evaluate four patch-wise DE models: three are depth refinement models including our PRO model, BoostingDepth \cite{boost}, PatchRefiner \cite{pr}, and the other one is a direct DE model that is PatchFusion \cite{pf}. For PatchFusion and PatchRefiner, we conduct evaluations under two settings: (i) `$\mathrm{P}$=16' setting that uses the same patch number as our method, and (ii) `$\mathrm{P}$=177' setting, where additional patches are used for test-time ensembling. Note that $\mathrm{P}$ denotes the number of patches used to reassemble final depth maps during inference. As shown in Table \ref{tab:main}, our PRO achieves SOTA performance across all depth metrics while maintaining the lowest inference time. Notably, our BFM prevents overfitting to synthetic dataset biases, leading to a 9.5\% improvement in AbsRel on the Booster dataset that contains transparent objects. In terms of edge quality, our method performs worse than BoostingDepth and PatchFusion in Boundary Recall (BR). However, as shown in the depth metrics, these two methods prioritize enhancing edge sharpness rather than focusing on enhancing overall depth accuracy. In contrast, when compared to PatchRefiner that exhibits similar depth performance, our PRO achieves superior results in terms of the edge metric (BR). To further demonstrate the generalizability of our method, we include additional experiments using alternative base models such as MiDaS v3.1 \cite{midav3} and DepthAnythingV1 \cite{da}, as detailed in the \emph{supplementary material}.

In terms of consistency, our PRO achieves the best result, with an 85.9\% improvement, as shown in Table \ref{tab:consistency}. It demonstrates the effectiveness of our GPCT strategy in maintaining consistency between individually processed patches.

\noindent \textbf{Qualitative Comparisons.} Fig.~\ref{fig:qual} shows qualitative comparisons for four patch-wise DE models. As shown in Fig.~\ref{fig:qual}, PatchFusion and PatchRefiner trained on the synthetic dataset without masking exhibit artifacts in the window regions. Additionally, two patch-based methods \cite{pr,pf} produce noticeable boundary artifacts when test ensembling is not applied. In contrast, despite refining each patch only once, our PRO demonstrates minimal depth discontinuity while achieving the fastest inference time.
\vspace{-0.2cm}

\enlargethispage{\baselineskip}
\subsection{Ablation Studies} \label{ab}
\vspace{-0.1cm}
We analyze the effectiveness of the core components of our PRO through ablation studies: Grouped Patch Consistency Training (GPCT) and Bias Free Masking (BFM).

\noindent \textbf{Ablation on GPCT and BFM.}  Table \ref{tab:ablation} presents the ablation results on GPCT and BFM. When BFM is applied to the baseline (a), the improvements are relatively minor for most datasets. However, for the Booster dataset \cite{booster} that contains transparent objects, we observe a significant improvement of 21.3\% in AbsRel metric, demonstrating that our BFM method effectively identifies unreliable regions. Even when only BFM is applied without GPCT, a reduction in consistency error (CE) suggests that supervision limited to reliable regions helps mitigate unnecessary refinement, thereby reducing confusion in the depth refinement process. Furthermore, applying GPCT to the baseline yields a 50.4\% improvement in consistency error, demonstrating its effectiveness in maintaining consistency between independently processed patches. Although model (c) achieves the best performance in the $\mathrm{D}^3\mathrm{R}$ metric, it introduces artifacts in transparent regions. Consequently, model (d), which incorporates both GPCT and BFM, achieves the best overall performance across most metrics.

\begin{table}[t]
    \scriptsize
    \centering
    \resizebox{0.3\textwidth}{!}{
    \def\arraystretch{1.2}
    \begin{tabular} {c|c|c|c}
        \Xhline{2\arrayrulewidth}
        \multirow{2}{*}{Overlap} & \multirow{2}{*}{CE$\downarrow$} & \textbf{ETH3D} & \textbf{Middle14} \\
        \cline{3-4}
        & & AbsRel$\downarrow$ & AbsRel$\downarrow$ \\
        \hline
        28 & 0.108 & 0.0423 & 0.0288 \\
        56 & 0.113 & 0.0424 & 0.0289 \\
        112 & 0.065 & 0.0425 & 0.0288 \\
        224 & \textbf{0.049} & \textbf{0.0422} & \textbf{0.0287} \\
        448 & 0.060 & 0.0423 & \textbf{0.0287} \\
       
        \Xhline{2\arrayrulewidth}
    \end{tabular}}
    \vspace{-0.2cm}
    \caption{Ablations on overlap sizes in Consistency Error (CE) for ETH3D \cite{eth} and Middlebury 2014 \cite{middle} datasets.}
    \vspace{-0.2cm}
    \label{tab:ablation_overlap}
\end{table}

\noindent \textbf{Ablation on Overlap Sizes in GPCT.} Table \ref{tab:ablation_overlap} shows the effect of different overlap sizes in GPCT. Each overlap value denotes the size (in pixels) of the square overlapping region shared between adjacent patches. As shown, increasing the overlap generally reduces the CE. However, when the overlap reaches 448 pixels, CE begins to increase. We attribute this to the fact that highly overlapped patches capture nearly identical scene contents, resulting in fewer inconsistencies between patches. So, the depth consistency loss provides less meaningful supervision, reducing its effectiveness.

\enlargethispage{\baselineskip}


\vspace{-0.3cm}

\section{Conclusion}
\label{sec:conclusion}
\vspace{-0.2cm}

In this paper, we propose the Patch Refine Once (PRO) model for depth refinement on high-resolution images. To address the depth discontinuity problem, we introduce the Grouped Patch Consistency Training (GPCT) strategy, ensuring consistency between independently processed patches. In addition, we propose Bias Free Masking (BFM) to prevent the depth refinement model from overfitting to dataset-specific biases. Through these strategies, our PRO achieves superior zero-shot performance while maintaining a low inference time (12$\times$ faster than PatchRefiner with 177 patches), outperforming recent state-of-the-art methods across diverse datasets.

\clearpage
\vspace{-1cm}
\noindent \textbf{Acknowledgements.}\quad This work was supported by IITP grant funded by the Korea government(MSIT) (No.RS2022-00144444, Deep Learning Based Visual Representational Learning and Rendering of Static and Dynamic Scenes).

\clearpage
\setcounter{page}{1}
\maketitlesupplementary
\appendix

\begin{strip}\centering
    \vspace{-1cm}

\scriptsize
\centering
\resizebox{0.9\textwidth}{!}{
\def\arraystretch{1.2}
\begin{tabular} {c|c|cc|cc|ccc|c|c}
    \Xhline{2\arrayrulewidth}
    \multirow{2}{*}{Base} & \multirow{2}{*}{Methods} & \multicolumn{2}{c|}{\textbf{Booster}} & \multicolumn{2}{c|}{\textbf{ETH3D}} & \multicolumn{3}{c|}{\textbf{Middle14}} & \textbf{DIS} & \multirow{2}{*}{CE $\downarrow$}\\
    \cline{3-10}
    & & AbsRel$\downarrow$ & $\delta_1\uparrow$ & AbsRel$\downarrow$ & $\delta_1\uparrow$ & AbsRel$\downarrow$ & $\delta_1\uparrow$ & $\mathrm{D}^3\mathrm{R}\downarrow$ & BR$\uparrow$ \\
    \hline
    \multirow{4}{*}{MiDaS v3.1\cite{midav3}} 
    & MiDaS v3.1 \cite{midav3} & 0.0538 & 0.973 & 0.0655 & 0.965 & 0.0399 & 0.991 & 0.1914 & 0.028 & - \\
    & PatchFusion $_{\mathrm{P}=16}$ \cite{pf} & 0.0613 & 0.964 & 0.0826 & 0.952 & 0.0465 & 0.989 & 0.1435 & \textbf{0.100} & 0.456\\
    & PatchRefiner $_{\mathrm{P}=16}$ \cite{pr}   & 0.0520 & \textbf{0.977} & 0.0594 & 0.971 & 0.0377 & \textbf{0.993} & 0.1396 & 0.058 & 0.461\\
    & \textbf{PRO}  & \textbf{0.0513} & \textbf{0.977} & \textbf{0.0578} & \textbf{0.972} & \textbf{0.0370} & \textbf{0.993} & \textbf{0.1261} & 0.062 & \textbf{0.077} \\
    \hline
    \multirow{4}{*}{DepthAnythingV1 \cite{da}} 
    & DepthAnythingV1 \cite{da} & \textbf{0.0482} & 0.977 & 0.0489 & 0.981 & 0.0315 & 0.995 & 0.1546 & 0.023 & -\\
    & PatchFusion $_{\mathrm{P}=16}$ \cite{pf} & 0.0621 & 0.962 & 0.0672 & 0.970 & 0.0410 & 0.993 & 0.1246 & \textbf{0.081} & 0.393 \\
    & PatchRefiner $_{\mathrm{P}=16}$ \cite{pr}  & 0.0507 & 0.977 & 0.0480 & 0.982 & 0.0301 & \textbf{0.996} & 0.1240 & 0.043 & 0.409\\
    & \textbf{PRO}  & 0.0500 & \textbf{0.978} & \textbf{0.0455} & \textbf{0.983} & \textbf{0.0294} & \textbf{0.996} & \textbf{0.1193} & 0.052 & \textbf{0.073} \\
    \Xhline{2\arrayrulewidth}
\end{tabular}}
\vspace{-0.2cm}
\captionof{table}{Generality across various base models. Quantitative comparison of depth estimation methods on Booster \cite{booster}, ETH3D \cite{eth}, Middlebury 2014 \cite{middle}, and DIS-5K \cite{dis} datasets. \textbf{Bold} indicates the best performance in each metric.}
\label{tab:othermodels}
    
\end{strip}

\begin{table}[t]
    \scriptsize
    \centering
    \resizebox{1.0\columnwidth}{!}{
    \def\arraystretch{1.2}
    \begin{tabular} {c|c|cc|cc|ccc|c}
        \Xhline{2\arrayrulewidth}
        \multirow{2}{*}{Meth} & \multirow{2}{*}{BFM} & \multicolumn{2}{c|}{\textbf{Booster}} & \multicolumn{2}{c|}{\textbf{ETH3D}} & \multicolumn{3}{c|}{\textbf{Middle14}} & \textbf{DIS} \\
        \cline{3-10}
        & & AbsRel$\downarrow$ & $\delta_1\uparrow$ & AbsRel$\downarrow$ & $\delta_1\uparrow$ & AbsRel$\downarrow$ & $\delta_1\uparrow$ & $\mathrm{D}^3\mathrm{R}\downarrow$ & BR$\uparrow$   \\
        \hline
        \multirow{2}{*}{PF [{\color{iccvblue}{23}}]} 
        &  & 0.0504 & 0.985 & 0.0735 & 0.956 & 0.0450 & 0.989 & 0.1124 & 0.206  \\
        & \checkmark & 0.0446 & 0.991 & 0.0761 & 0.956 & 0.0447 & 0.991 & 0.1149 & 0.198 \\
        \hline
        \multirow{2}{*}{PR [{\color{iccvblue}{24}}]} 
        &  & 0.0348 & 0.989 & 0.0435 & 0.985 & 0.0292 & 0.995 & 0.0830  & 0.151  \\
        & \checkmark & 0.0307 & 0.994 & 0.0432 & 0.985 & 0.0290 & 0.995 & 0.0791 & 0.151  \\
        \Xhline{2\arrayrulewidth}
    \end{tabular}}
    \caption{Effect of Bias Free Masking (BFM) on training of PatchFusion (PF) \cite{pf} and PatchRefiner (PR) \cite{pr}. }
    \label{tab:bfm}
\end{table}

\section{Extra Experiments}

\begin{figure}[htbp]
  \centering
  \includegraphics[width=1.0\columnwidth]{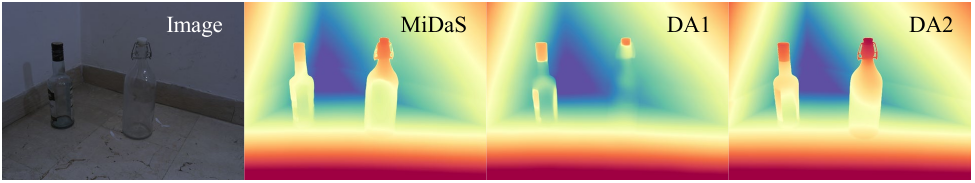}
  \vspace{-0.5cm}
  \caption{Depth estimation results from different base models (MiDaS v3.1 \cite{midav3}, DepthAnythingV1 \cite{da}, and DepthanythingV2 \cite{da2}.}
  \label{fig:trans}
\end{figure}

\noindent \textbf{Generality across Base Models.} Replacing the base model (BM) requires retraining. To show the generality of our method, we provide additional comparisons using MiDaS v3.1 \cite{midav3} and DepthAnythingV1 (DA1) \cite{da} as base models. All evaluations are conducted in the same way in the Section~\ref{perf_comp}. As shown in the Table \ref{tab:othermodels}, the MiDaS v3.1- and DA1-based PRO consistently outperform the previous SOTA models. However, all DA1-based methods show inferior AbsRel on the Booster compared to the  DA1. This is because, as shown in the Fig.~\ref{fig:trans}, DA1 yields poor predictions on transparent objects, thus leading the other methods to further degraded performance.

\noindent \textbf{Effect of Bias Free Masking (BFM).} To further demonstrate the effectiveness of BFM, we apply it to both PatchFusion \cite{pf} and PatchRefiner \cite{pr} during training. The consistent performance improvements observed across most metrics validate the generalizability of our approach, as summarized in Table \ref{tab:bfm}.

\begin{table}[t]
    \scriptsize
    \centering
    \resizebox{1.0\columnwidth}{!}{
    \def\arraystretch{1.2}
    \begin{tabular} {c|cc|cc|ccc|c}
        \Xhline{2\arrayrulewidth}
        \multirow{2}{*}{P} & \multicolumn{2}{c|}{\textbf{Booster}} & \multicolumn{2}{c|}{\textbf{ETH3D}} & \multicolumn{3}{c|}{\textbf{Middle14}} & \textbf{DIS}  \\
        \cline{2-9}
         & AbsRel$\downarrow$ & $\delta_1\uparrow$ & AbsRel$\downarrow$ & $\delta_1\uparrow$ & AbsRel$\downarrow$ & $\delta_1\uparrow$ & $\mathrm{D}^3\mathrm{R}\downarrow$ & BR$\uparrow$ \\
        \hline
        $3 \times 3$  & 0.0305 & 0.994 & 0.0423 & 0.985 & 0.0288 & 0.995 & 0.0805 & 0.125 \\
        $4 \times 4$  & 0.0304 & 0.994 & 0.0422 & 0.985 & 0.0287 & 0.996 & 0.0803 & 0.156 \\
        $5 \times 5$  & 0.0305 & 0.994 & 0.0423 & 0.985 & 0.0288 & 0.996 & 0.0797 & 0.162  \\
        \Xhline{2\arrayrulewidth}
    \end{tabular}}
    \caption{Effect of patch numbers on the performance. }
    \label{tab:patches}
\end{table}

\begin{table}[t]
    \scriptsize
    \centering
    \resizebox{1.0\columnwidth}{!}{
    \def\arraystretch{1.2}
    \begin{tabular} {c|c|cc|cc|ccc|c}
        \Xhline{2\arrayrulewidth}
        \multirow{2}{*}{Base} & \multirow{2}{*}{CE $\downarrow$} & \multicolumn{2}{c|}{\textbf{Booster}} & \multicolumn{2}{c|}{\textbf{ETH3D}} & \multicolumn{3}{c|}{\textbf{Middle14}} & \textbf{DIS} \\
        \cline{3-10}
        & & AbsRel$\downarrow$ & $\delta_1\uparrow$ & AbsRel$\downarrow$ & $\delta_1\uparrow$ & AbsRel$\downarrow$ & $\delta_1\uparrow$ & $\mathrm{D}^3\mathrm{R}\downarrow$ & BR$\uparrow$ \\
        \hline
        Diag & 0.075 & 0.0302 & 0.994 & 0.0423 & 0.986 & 0.0288 & 0.996 & 0.0787 & 0.171 \\
        GPCT & 0.049 & 0.0304 & 0.994 & 0.0422 & 0.985 & 0.0287 & 0.996 & 0.0803 & 0.156\\
        \Xhline{2\arrayrulewidth}
    \end{tabular}}
    \caption{Quantitative comparisons between regularizing consistency diagonally (Diag) and GPCT.}
    \label{tab:gpct}
\end{table}

\noindent \textbf{Effect of Patch numbers on Performance.} To investigate the effect of the number of patches during inference, we conduct experiments with the PRO model by partitioning the input images into 3$\times$3, 4$\times$4, and 5$\times$5 patches, and compare their performance.
As we can see in the Table \ref{tab:patches}, depth metrics remain stable across patch numbers, while edge metrics improve with more patches. This allows users to adjust patch counts at inference to balance speed and detail without retraining.

\begin{figure}[h]
  \centering
  \includegraphics[width=1.0\columnwidth]{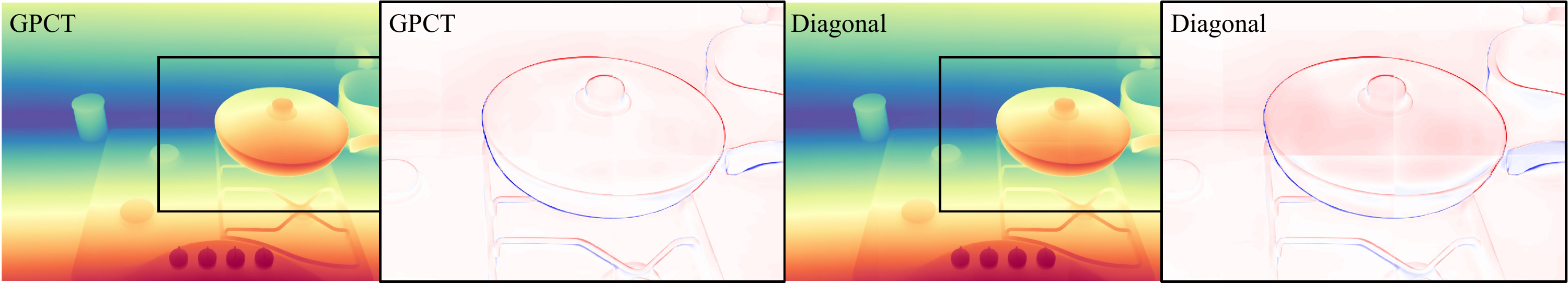}
  \caption{Qualitative comparisons between Diagonal approach and GPCT. }
  \label{fig:diag}
\end{figure}

\noindent \textbf{Ablation on GPCT vs Diagonal consistency.} We additionally provide comparison between regularizing 2-diagonal patches and our GPCT in the Table \ref{tab:gpct}. While the depth metrics show marginal differences, GPCT achieves significantly better consistency in CE metric qualitatively as shown in the Fig.\ref{fig:diag}.

\begin{table*}\centering
    \includegraphics[width=1.0\linewidth]{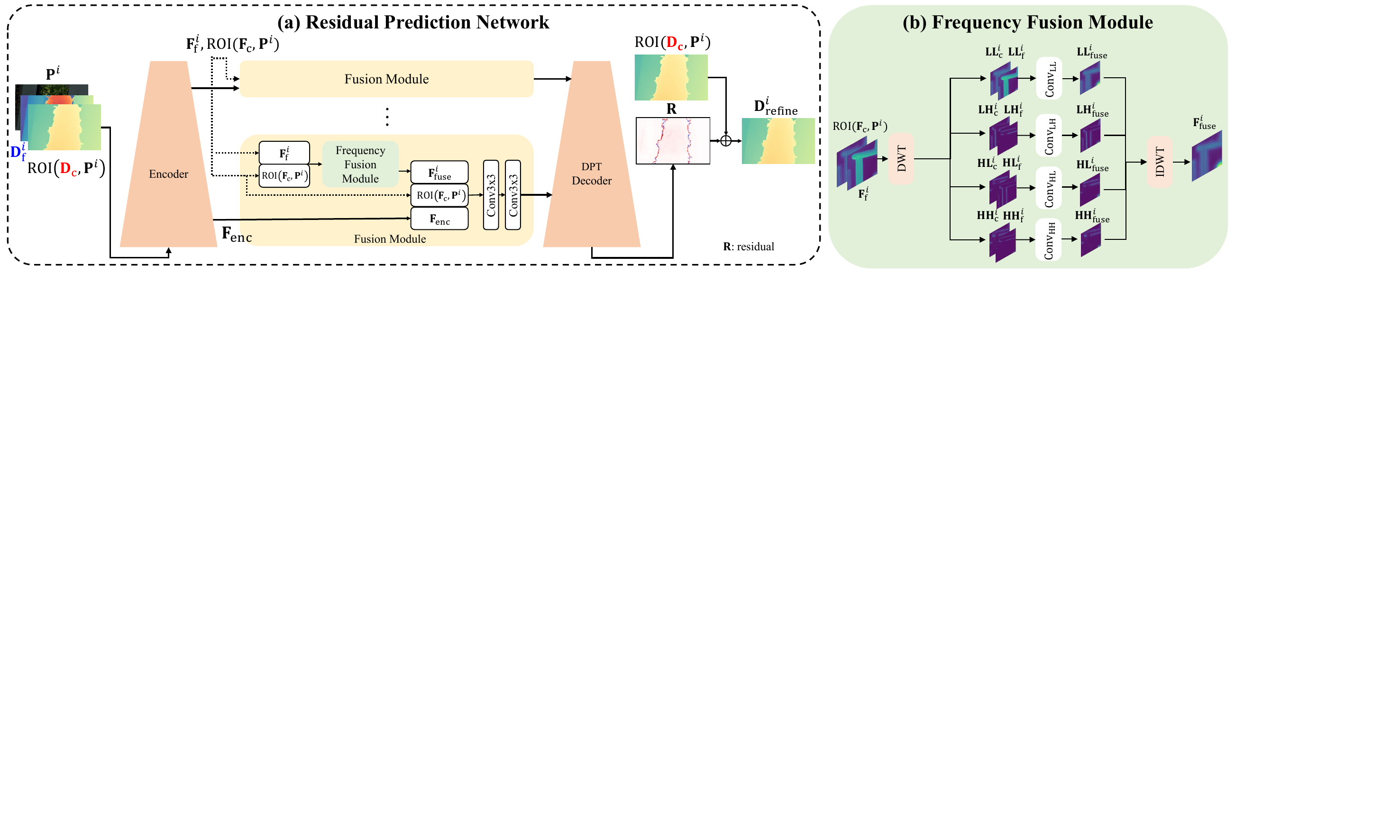}
    \vspace{-0.6cm}
    \captionof{figure}{\textbf{Architecture of the Residual Prediction Network and Frequency Fusion Module (FFM).} \textbf{(a) Residual Prediction Network.} The Residual Prediction Network comprises an encoder, a decoder, and the Fusion Module. \textbf{(b) Frequency Fusion Module (FFM).} We utilize Discrete Wavelet Transform (DWT) to decompose the input features into four frequency components. Then, each frequency component is processed independently using convolutional layers.}
    \label{fig:supple_main}
    
\end{table*}

\begin{table*}[t]
    \scriptsize
    \centering
    \resizebox{1.0\textwidth}{!}{
    \def\arraystretch{1.2}
    \begin{tabular} {l|c|c|c|c|ccc|cc|cc|cc}
        \Xhline{2\arrayrulewidth}
        \multirow{2}{*}{Models} & \multirow{2}{*}{\makecell{FLOPs \\ (Mac)}} & \multirow{2}{*}{Parameter} 
        & \textbf{DIS}  & \textbf{UHRSD} & \multicolumn{3}{c|}{\textbf{Middle14}} & \multicolumn{2}{c|}{\textbf{ETH3D}} & \multicolumn{2}{c|}{\textbf{Booster}} & \multicolumn{2}{c}{\textbf{NuScenes}}\\
        \cline{4-14}
        & & & BR$\uparrow$ & BR$\uparrow$ & AbsRel$\downarrow$ & $\delta_1\uparrow$ & $\mathrm{D}^3\mathrm{R}\downarrow$ & AbsRel$\downarrow$ & $\delta_1\uparrow$ & AbsRel$\downarrow$ & $\delta_1\uparrow$ & AbsRel$\downarrow$ & $\delta_1\uparrow$\\
        \hline
        Conv & 31664G & 63M & 0.127 & 0.071 & 0.0290 & 0.995 & 0.0842 & 0.0428 & 0.984 & 0.0306 & 0.993 & 0.106 & 0.882 \\
        \textbf{FFM (Ours)} & 20250G & 51M & \textbf{0.156} & \textbf{0.083} & \textbf{0.0287} & \textbf{0.996} & \textbf{0.0803} & \textbf{0.0422} & \textbf{0.985} & \textbf{0.0304} & \textbf{0.994} & \textbf{0.104} & \textbf{0.883} \\
        \Xhline{2\arrayrulewidth}
    \end{tabular}}
    \vspace{-0.2cm}
    \caption{Ablation study of the Frequency Fusion Module (FFM) on DIS-5K \cite{dis}, UHRSD \cite{uhrsd}, Middlebury 2014 \cite{middle}, ETH3D \cite{eth}, Booster \cite{booster}, and NuScenes \cite{nuscenes}. \textbf{Bold} indicates the best performance in each metric.}
    \vspace{-0.3cm}
    \label{tab:ab_ffm}
\end{table*}


\section{Architecture of the Fusion Module}
The encoder takes $\mathbf{P}^{i}$, $\mathbf{D}_{\mathrm{f}}^{i}$, and $\mathsf{ROI}(\mathbf{D}_{\mathrm{c}}, \mathbf{P}^{i})$ as inputs and produces a set of five intermediate features, denoted as $\mathbf{F}_{\mathrm{enc}} = \{ \mathbf{f}_{\mathrm{enc}, j}^{i} \}_{j=1}^{5}$ following \cite{pf}. Then, the fused feature map $\mathbf{F}_{\mathrm{fuse}}^{i}$ is obtained through the frequency fusion module ($\mathsf{FFM}$), defined as 
$
\mathbf{F}_{\mathrm{fuse}}^{i} = \mathsf{FFM}(\mathsf{concat}(\mathbf{F}_{\mathrm{f}}^{i}, \mathsf{ROI}(\mathbf{F}_{\mathrm{c}}, \mathbf{P}^{i}))).
$
Subsequently, $\mathsf{concat}(\mathbf{F}_{\mathrm{fuse}}^{i}, \mathsf{ROI}(\mathbf{F}_{\mathrm{c}},\mathbf{P}^{i}), \mathbf{F}_{\mathrm{enc}})$ is processed through two consecutive layers, each consisting of a $3 \times 3$ convolution, batch normalization, and ReLU activation. Finally, the resulting feature is fed into the DPT decoder \cite{dpt} to obtain the residual map $\mathbf{R}$.

\noindent\textbf{Architecture of the Frequency Fusion Module (FFM)}
To obtain accurate depth values from the coarse depth and preserve fine details from the fine depth, we design a Frequency Fusion Module (FFM) that effectively extracts and integrates edge information. We utilize Discrete Wavelet Transform (DWT) to decompose the input features into four frequency components: LL, LH, HL, and HH, which represent the low-frequency and high-frequency information. Each component is processed with its own dedicated convolution to capture scale-specific features. Finally, the components are recombined using the Inverse Discrete Wavelet Transform (IDWT), resulting in fused features that retain both global depth consistency and enhanced edge details. Overall process is described in Fig.~\ref{fig:supple_main}-(b). To describe this process in more detail, we first decompose $\mathsf{ROI}(\mathbf{F}_{\mathrm{c}}, \mathbf{P}^{i})$ and $\mathbf{F}_{\mathrm{f}}^{i}$ into four frequency sub-bands $\mathbf{X}$ ($\forall \mathbf{X} \in \{\mathbf{LL}, \mathbf{LH}, \mathbf{HL}, \mathbf{HH}\}$) using $\mathsf{DWT}$. Each sub-band is then fused using a corresponding convolution $\text{Conv}_{\mathbf{X}}$. Finally, the fused feature map $\mathbf{F}_{\mathrm{fuse}}^{i}$ is obtained through $\mathsf{IDWT}$.

\begin{equation}
\mathbf{X}_{\mathrm{c}}^{i}, \mathbf{X}_{\mathrm{f}}^{i} = \mathsf{DWT}(\mathsf{ROI}(\mathbf{F}_{\mathrm{c}}, \mathbf{P}^{i})), \mathsf{DWT}(\mathbf{F}_{\mathrm{f}}^{i}) 
\end{equation}

\begin{equation}
\mathbf{X}_{\text{fuse}}^{i} = \text{Conv}_{\mathbf{X}} (\mathsf{concat}(\mathbf{X}_{\mathrm{c}}^{i}, \mathbf{X}_{\mathrm{f}}^{i}))
\end{equation}

\begin{equation}
\mathbf{F}_{\text{fuse}}^{i} = \mathsf{IDWT}(\mathbf{LL}_{\text{fuse}}^{i}, \mathbf{LH}_{\text{fuse}}^{i}, \mathbf{HL}_{\text{fuse}}^{i}, \mathbf{HH}_{\text{fuse}}^{i})
\end{equation}

\subsection{Ablation study of FFM}

To validate the effectiveness of the Frequency Fusion Module (FFM), we conduct an ablation study by replacing the FFM with a simple convolutional block consisting of Conv-ReLU-Conv layers. To ensure that any performance gain is not simply due to an increase in the number of parameters or FLOPs, we design the simple convolutional block to have more parameters and FLOPs than the FFM. This allows us to attribute the performance improvement to the design of FFM itself, rather than computational complexity. As shown in the Table \ref{tab:ab_ffm}, our method not only achieves the best performance in standard depth metrics, but also yields significant improvements in edge accuracy. Specifically, it achieves a 22.5\% improvement on the DIS-5K dataset and a 16.9\% improvement on the UHRSD dataset in the Boundary Recall (BR) metric. In addition, we observe a 4.6\% improvement in the edge quality metric ($\mathrm{D}^3\mathrm{R}$). It demonstrates that the proposed FFM effectively integrates edge information through the use of Discrete Wavelet Transform (DWT), which enables selective enhancement of high-frequency details without sacrificing global structure. This highlights the benefit of frequency-domain processing in depth refinement tasks.

\section{Additional Implementation Details}

\noindent \textbf{Details in training.} Before training, we first filter out unnecessary training samples based on the unreliable mask, $\textbf{M}_{\mathrm{unreliable}}$, described in Section~\ref{mask}. Specifically, we load each $\textbf{M}_{\mathrm{unreliable}}$ using NumPy and discard the corresponding training sample if the mean value of $\textbf{M}_{\mathrm{unreliable}}$ exceeds 0.5, indicating excessive unreliable regions. During training, we randomly crop a region (e.g., 846$\times$1505) from a 2160$\times$3840 input to generate 2$\times$2 overlapping patches (540$\times$960 each), which are resized to 518$\times$518 for refinement. The overlap size (234$\times$415) corresponds to $\frac{224}{518}\approx 43\%$ of the patch dimensions, where 224 is the empirically chosen overlap value from Section~\ref{ab}.

\noindent \textbf{Details in inference.} At inference, the input is divided into a 4$\times$4 non-overlapping grid, and the refined 518$\times$518 patches are reassembled into a 2072$\times$2072 depth map. The final depth map is then upsampled to the original resolution via bilinear interpolation. This inference procedure is consistent with PF$_{\mathrm{P}=16}$ and PR$_{\mathrm{P}=16}$. As demonstrated in Table \ref{tab:patches}, the number of patches can be flexibly adjusted depending on the inference setting.

\section{Qualitative Results}
\begin{figure*}
    \centerline{\includegraphics[width=0.9\textwidth]{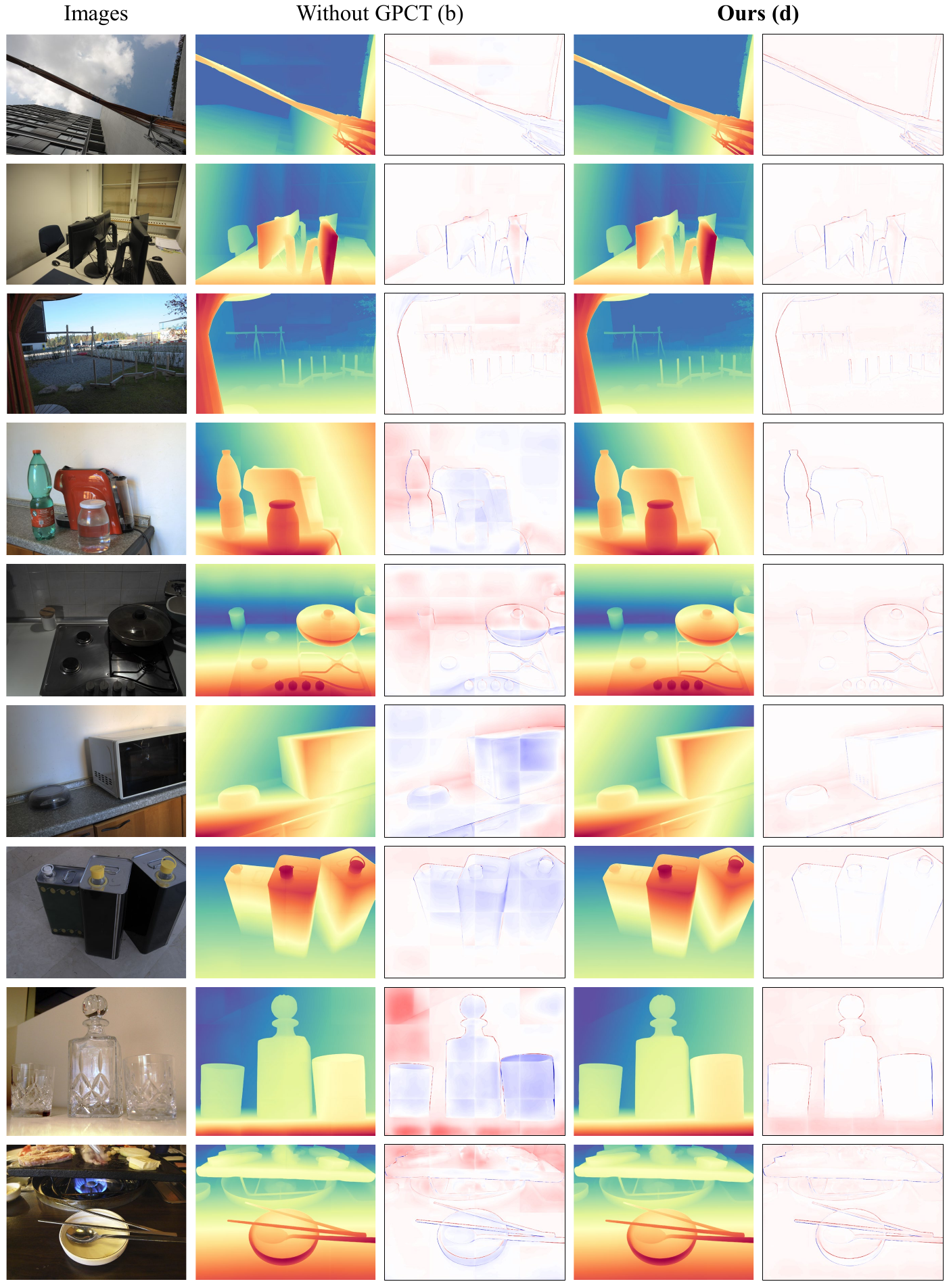}}
    \caption{\textbf{Qualitative comparisons of GPCT's impact on ETH3D \cite{eth}, Booster \cite{booster}, and DIS-5k \cite{dis}.} We compare PRO (Ours (d)) with the model trained without Grouped Patch Consistency Training (GPCT) (b). (b) and (d) represent the model index in Table \ref{tab:ablation}. We also visualize the residuals to highlight the presence of artifacts more effectively. Zoom in for details.} 
    \label{fig:ab_gpct}
\end{figure*}

\begin{figure*}
    \centerline{\includegraphics[width=1.0\textwidth]{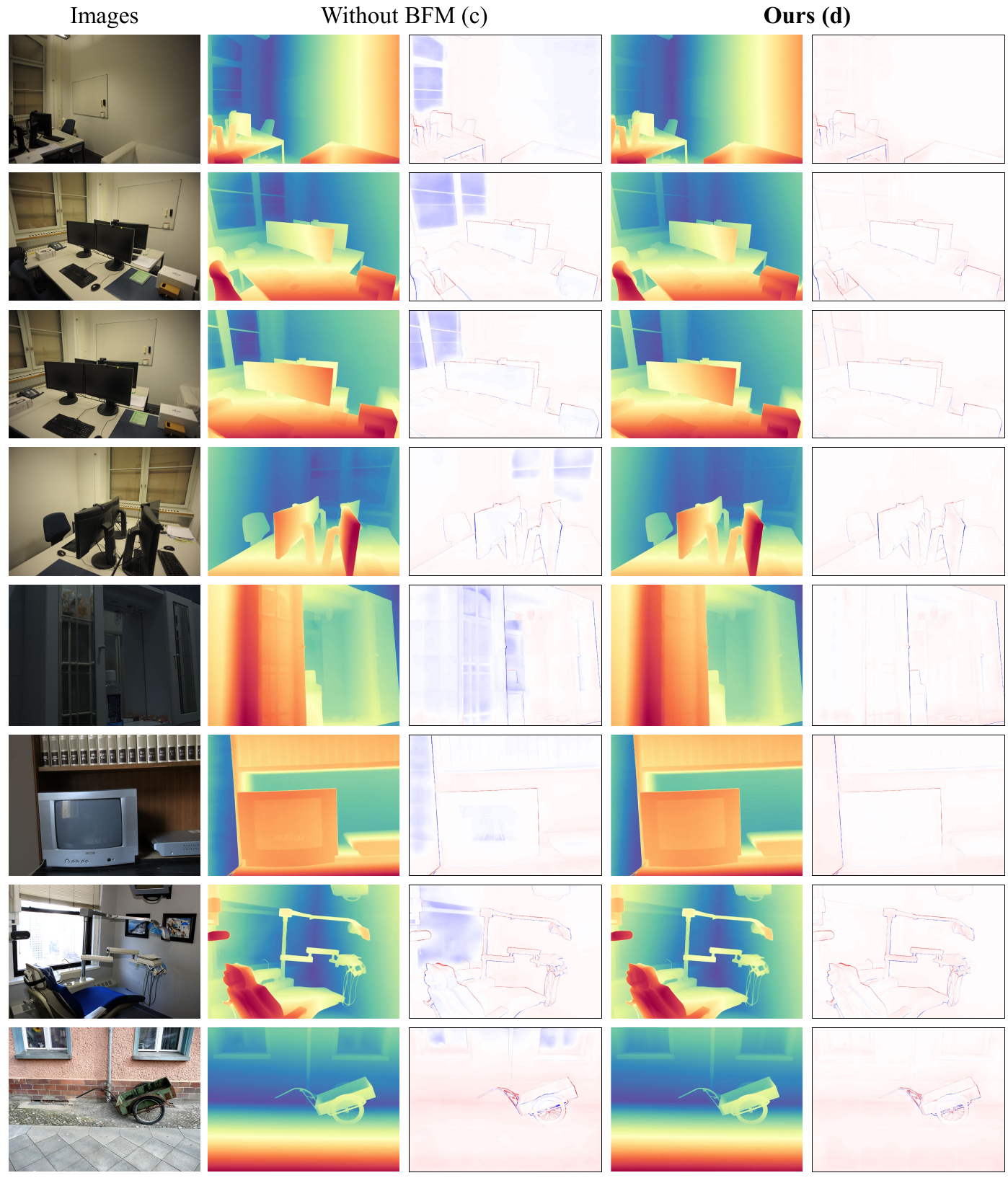}}
    \caption{\textbf{Qualitative comparisons of BFM's impact on ETH3D \cite{eth}, Booster \cite{booster}, and DIS-5k \cite{dis}.} We compare PRO (Ours (d)) with the model trained without Bias Free Masking (BFM) (c). (c) and (d) represent the model index in Table \ref{tab:ablation}. We also visualize the residuals to highlight the presence of artifacts more effectively. Zoom in for details.} 
    \label{fig:ab_bfm}
\end{figure*}

\noindent \textbf{Ablation Study}
In the ablation study (Section \ref{ab}), we analyze the effect of Grouped Patch Consistency Training (GPCT) and Bias Free Masking (BFM) quantitatively. In this section, we analyze the effect of GPCT and BFM with qualitative results. As shown in the Fig.~\ref{fig:ab_gpct}, the model trained without GPCT shows remarkable depth discontinuity problem on the grids. On the other hand, our PRO model trained with GPCT alleviates the depth discontinuity problem. Likewise, as shown in the Fig.~\ref{fig:ab_bfm} , the model trained without BFM exhibits artifacts on transparent surfaces, such as glass windows, as well as reflective surfaces like TV screens. In contrast, our model trained with BFM effectively refines only the edge regions while preventing artifacts on transparent objects.

\begin{figure*}
    \centerline{\includegraphics[width=1.0\textwidth]{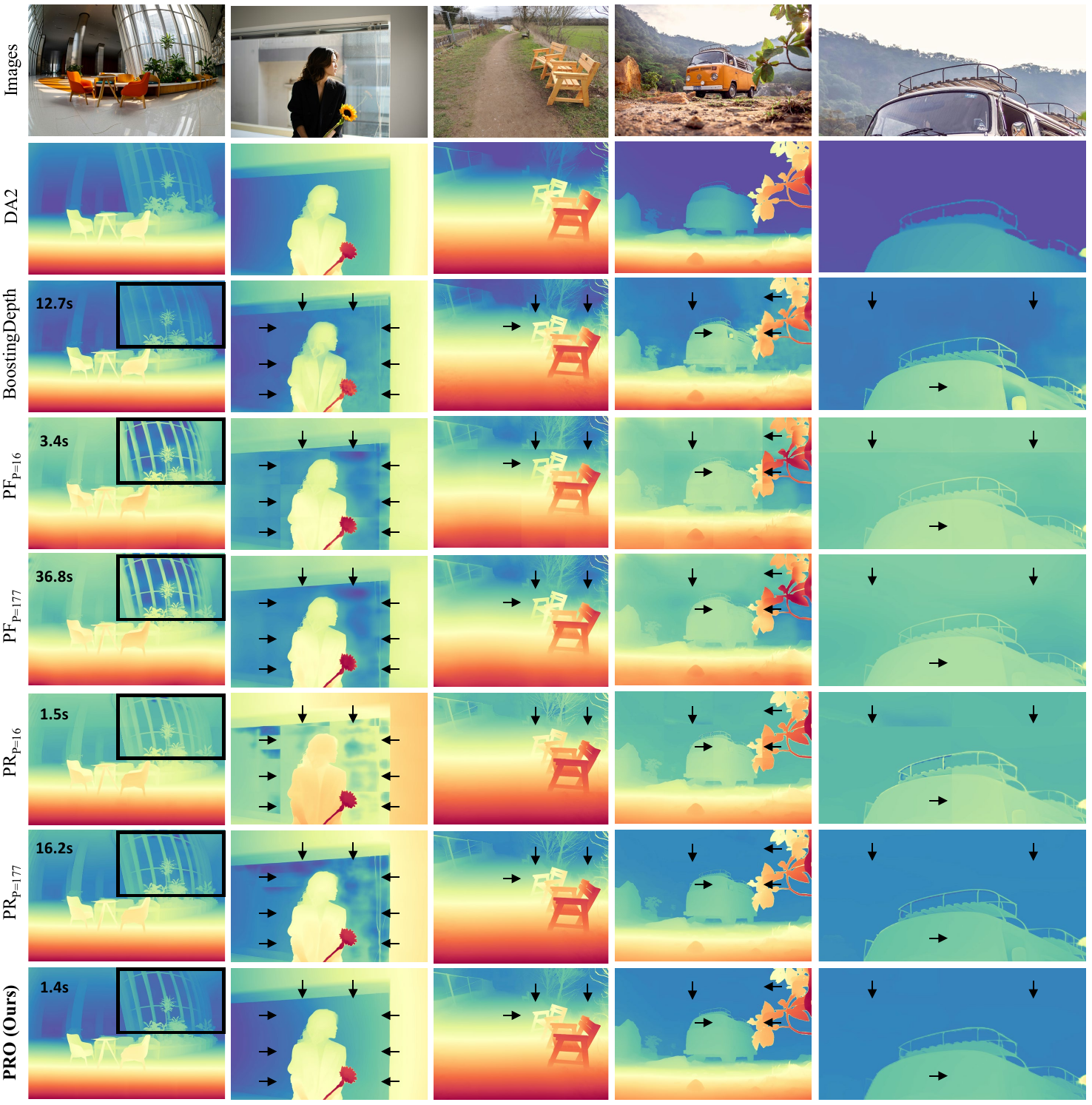}}
    \caption{\textbf{Qualitative comparisons for patch-wise DE methods on UHRSD \cite{uhrsd} and images from the internet.} We compare PRO (Ours) with BoostingDepth \cite{boost}, PatchFusion (PF) \cite{pf}, and PatchRefiner (PR) \cite{pr}. The time displayed in the leftmost depth column represents the inference time. Black rectangle area represents the transparent object region. Black arrows indicate patch boundaries. Zoom in for details.} 
    \label{fig:qual_supple}
\end{figure*}

\begin{figure*}
    \centerline{\includegraphics[width=1.0\textwidth]{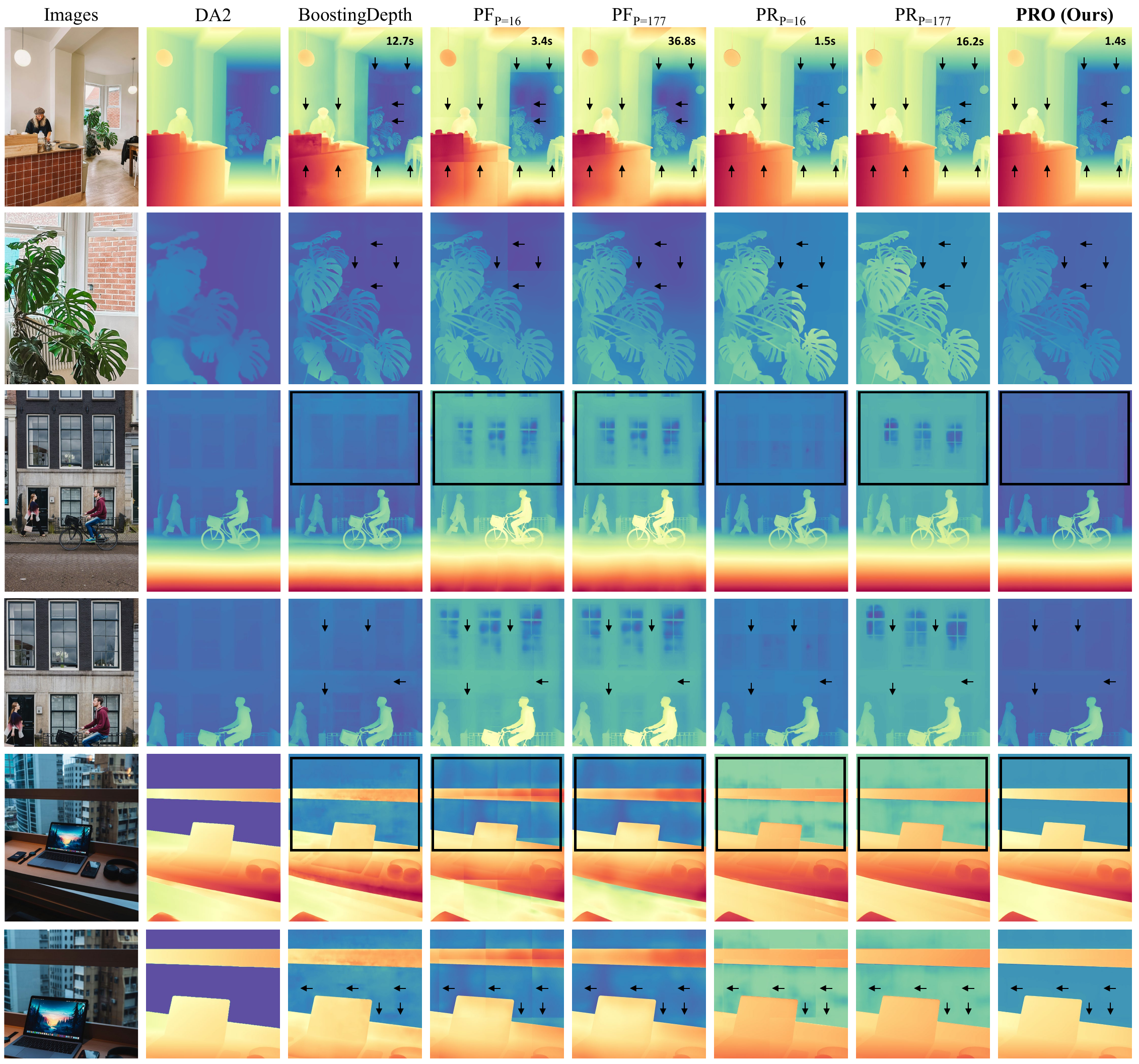}}
    \caption{\textbf{Qualitative comparisons for patch-wise DE methods on images from the internet.} We compare PRO (Ours) with BoostingDepth \cite{boost}, PatchFusion (PF) \cite{pf}, and PatchRefiner (PR) \cite{pr}. The time displayed in the top depth row represents the inference time. Black rectangle area represents the transparent object region. Black arrows indicate patch boundaries. Zoom in for details.} 
    \label{fig:qual_supple_2}
\end{figure*}

\noindent \textbf{Additional Qualitative Results}
We provide additional qualitative comparisons of BoostingDepth \cite{boost}, PatchFusion \cite{pf}, PatchRefiner \cite{pr}, and PRO (Ours) on the UHRSD \cite{uhrsd} dataset and on internet images (e.g., from Unsplash\footnote{\url{https://unsplash.com}} and Pexels\footnote{\url{https://www.pexels.com}}), as shown in Fig.\ref{fig:qual_supple} and Fig.\ref{fig:qual_supple_2}.

\clearpage
{
    \small
    \bibliographystyle{ieeenat_fullname}
    \bibliography{reference}

\begin{thebibliography}{49}
\providecommand{\natexlab}[1]{#1}
\providecommand{\url}[1]{\texttt{#1}}
\expandafter\ifx\csname urlstyle\endcsname\relax
  \providecommand{\doi}[1]{doi: #1}\else
  \providecommand{\doi}{doi: \begingroup \urlstyle{rm}\Url}\fi

\bibitem[Bhat et~al.(2021)Bhat, Alhashim, and Wonka]{adabins}
Shariq~Farooq Bhat, Ibraheem Alhashim, and Peter Wonka.
\newblock Adabins: Depth estimation using adaptive bins.
\newblock In \emph{Proceedings of the IEEE/CVF conference on computer vision and pattern recognition}, pages 4009--4018, 2021.

\bibitem[Birkl et~al.(2023)Birkl, Wofk, and M{\"u}ller]{midav3}
Reiner Birkl, Diana Wofk, and Matthias M{\"u}ller.
\newblock Midas v3. 1--a model zoo for robust monocular relative depth estimation.
\newblock \emph{arXiv preprint arXiv:2307.14460}, 2023.

\bibitem[Bochkovskii et~al.(2024)Bochkovskii, Delaunoy, Germain, Santos, Zhou, Richter, and Koltun]{pro}
Aleksei Bochkovskii, Ama{\"e}l Delaunoy, Hugo Germain, Marcel Santos, Yichao Zhou, Stephan~R Richter, and Vladlen Koltun.
\newblock Depth pro: Sharp monocular metric depth in less than a second.
\newblock \emph{arXiv preprint arXiv:2410.02073}, 2024.

\bibitem[Caesar et~al.(2020)Caesar, Bankiti, Lang, Vora, Liong, Xu, Krishnan, Pan, Baldan, and Beijbom]{nuscenes}
Holger Caesar, Varun Bankiti, Alex~H Lang, Sourabh Vora, Venice~Erin Liong, Qiang Xu, Anush Krishnan, Yu Pan, Giancarlo Baldan, and Oscar Beijbom.
\newblock nuscenes: A multimodal dataset for autonomous driving.
\newblock In \emph{Proceedings of the IEEE/CVF conference on computer vision and pattern recognition}, pages 11621--11631, 2020.

\bibitem[Chen et~al.(2021)Chen, Liu, and Wang]{inr}
Yinbo Chen, Sifei Liu, and Xiaolong Wang.
\newblock Learning continuous image representation with local implicit image function.
\newblock In \emph{Proceedings of the IEEE/CVF conference on computer vision and pattern recognition}, pages 8628--8638, 2021.

\bibitem[Cordts et~al.(2016)Cordts, Omran, Ramos, Rehfeld, Enzweiler, Benenson, Franke, Roth, and Schiele]{Cityscapes}
Marius Cordts, Mohamed Omran, Sebastian Ramos, Timo Rehfeld, Markus Enzweiler, Rodrigo Benenson, Uwe Franke, Stefan Roth, and Bernt Schiele.
\newblock The cityscapes dataset for semantic urban scene understanding.
\newblock In \emph{Proc. of the IEEE Conference on Computer Vision and Pattern Recognition (CVPR)}, 2016.

\bibitem[Dai et~al.(2023)Dai, Yi, Zhu, He, and Xu]{gbdf}
Yaqiao Dai, Renjiao Yi, Chenyang Zhu, Hongjun He, and Kai Xu.
\newblock Multi-resolution monocular depth map fusion by self-supervised gradient-based composition.
\newblock In \emph{Proceedings of the AAAI Conference on Artificial Intelligence}, pages 488--496, 2023.

\bibitem[Daubechies(1990)]{daubechies1990wavelet}
Ingrid Daubechies.
\newblock The wavelet transform, time-frequency localization and signal analysis.
\newblock \emph{IEEE transactions on information theory}, 36\penalty0 (5):\penalty0 961--1005, 1990.

\bibitem[Eftekhar et~al.(2021)Eftekhar, Sax, Malik, and Zamir]{omni}
Ainaz Eftekhar, Alexander Sax, Jitendra Malik, and Amir Zamir.
\newblock Omnidata: A scalable pipeline for making multi-task mid-level vision datasets from 3d scans.
\newblock In \emph{Proceedings of the IEEE/CVF International Conference on Computer Vision}, pages 10786--10796, 2021.

\bibitem[Eigen et~al.(2014)Eigen, Puhrsch, and Fergus]{eigen2014depth}
David Eigen, Christian Puhrsch, and Rob Fergus.
\newblock Depth map prediction from a single image using a multi-scale deep network.
\newblock \emph{Advances in neural information processing systems}, 27, 2014.

\bibitem[Fu et~al.(2024)Fu, Yin, Hu, Wang, Ma, Tan, Shen, Lin, and Long]{geowizard}
Xiao Fu, Wei Yin, Mu Hu, Kaixuan Wang, Yuexin Ma, Ping Tan, Shaojie Shen, Dahua Lin, and Xiaoxiao Long.
\newblock Geowizard: Unleashing the diffusion priors for 3d geometry estimation from a single image.
\newblock In \emph{European Conference on Computer Vision}, pages 241--258. Springer, 2024.

\bibitem[Geiger et~al.(2013)Geiger, Lenz, Stiller, and Urtasun]{kitti}
Andreas Geiger, Philip Lenz, Christoph Stiller, and Raquel Urtasun.
\newblock Vision meets robotics: The kitti dataset.
\newblock \emph{The international journal of robotics research}, 32\penalty0 (11):\penalty0 1231--1237, 2013.

\bibitem[Godard et~al.(2017)Godard, Mac~Aodha, and Brostow]{md1}
Cl{\'e}ment Godard, Oisin Mac~Aodha, and Gabriel~J Brostow.
\newblock Unsupervised monocular depth estimation with left-right consistency.
\newblock In \emph{Proceedings of the IEEE conference on computer vision and pattern recognition}, pages 270--279, 2017.

\bibitem[Godard et~al.(2019)Godard, Mac~Aodha, Firman, and Brostow]{md2}
Cl{\'e}ment Godard, Oisin Mac~Aodha, Michael Firman, and Gabriel~J Brostow.
\newblock Digging into self-supervised monocular depth estimation.
\newblock In \emph{Proceedings of the IEEE/CVF international conference on computer vision}, pages 3828--3838, 2019.

\bibitem[Guizilini et~al.(2020)Guizilini, Ambrus, Pillai, Raventos, and Gaidon]{guizilini20203d}
Vitor Guizilini, Rares Ambrus, Sudeep Pillai, Allan Raventos, and Adrien Gaidon.
\newblock 3d packing for self-supervised monocular depth estimation.
\newblock In \emph{Proceedings of the IEEE/CVF conference on computer vision and pattern recognition}, pages 2485--2494, 2020.

\bibitem[He et~al.(2017)He, Gkioxari, Doll{\'a}r, and Girshick]{maskrcnn}
Kaiming He, Georgia Gkioxari, Piotr Doll{\'a}r, and Ross Girshick.
\newblock Mask r-cnn.
\newblock In \emph{Proceedings of the IEEE international conference on computer vision}, pages 2961--2969, 2017.

\bibitem[Hua et~al.(2020)Hua, Kohli, Uplavikar, Ravi, Gunaseelan, Orozco, and Li]{holopix}
Yiwen Hua, Puneet Kohli, Pritish Uplavikar, Anand Ravi, Saravana Gunaseelan, Jason Orozco, and Edward Li.
\newblock Holopix50k: A large-scale in-the-wild stereo image dataset.
\newblock \emph{arXiv preprint arXiv:2003.11172}, 2020.

\bibitem[Huang et~al.(2018)Huang, Matzen, Kopf, Ahuja, and Huang]{mvssynth}
Po-Han Huang, Kevin Matzen, Johannes Kopf, Narendra Ahuja, and Jia-Bin Huang.
\newblock Deepmvs: Learning multi-view stereopsis.
\newblock In \emph{Proceedings of the IEEE conference on computer vision and pattern recognition}, pages 2821--2830, 2018.

\bibitem[Ke et~al.(2024)Ke, Obukhov, Huang, Metzger, Daudt, and Schindler]{marigold}
Bingxin Ke, Anton Obukhov, Shengyu Huang, Nando Metzger, Rodrigo~Caye Daudt, and Konrad Schindler.
\newblock Repurposing diffusion-based image generators for monocular depth estimation.
\newblock In \emph{Proceedings of the IEEE/CVF Conference on Computer Vision and Pattern Recognition}, pages 9492--9502, 2024.

\bibitem[Koch et~al.(2018)Koch, Liebel, Fraundorfer, and Korner]{ibims}
Tobias Koch, Lukas Liebel, Friedrich Fraundorfer, and Marco Korner.
\newblock Evaluation of cnn-based single-image depth estimation methods.
\newblock In \emph{Proceedings of the European Conference on Computer Vision (ECCV) Workshops}, pages 0--0, 2018.

\bibitem[Li et~al.(2025)Li, Wang, Zheng, Huang, Xian, Cao, and Zhang]{sddr}
Jiaqi Li, Yiran Wang, Jinghong Zheng, Zihao Huang, Ke Xian, Zhiguo Cao, and Jianming Zhang.
\newblock Self-distilled depth refinement with noisy poisson fusion.
\newblock \emph{Advances in Neural Information Processing Systems}, 37:\penalty0 69999--70025, 2025.

\bibitem[Li and Snavely(2018)]{mega}
Zhengqi Li and Noah Snavely.
\newblock Megadepth: Learning single-view depth prediction from internet photos.
\newblock In \emph{Proceedings of the IEEE conference on computer vision and pattern recognition}, pages 2041--2050, 2018.

\bibitem[Li et~al.(2024{\natexlab{a}})Li, Bhat, and Wonka]{pf}
Zhenyu Li, Shariq~Farooq Bhat, and Peter Wonka.
\newblock Patchfusion: An end-to-end tile-based framework for high-resolution monocular metric depth estimation.
\newblock In \emph{Proceedings of the IEEE/CVF Conference on Computer Vision and Pattern Recognition}, pages 10016--10025, 2024{\natexlab{a}}.

\bibitem[Li et~al.(2024{\natexlab{b}})Li, Bhat, and Wonka]{pr}
Zhenyu Li, Shariq~Farooq Bhat, and Peter Wonka.
\newblock Patchrefiner: Leveraging synthetic data for real-domain high-resolution monocular metric depth estimation.
\newblock In \emph{European Conference on Computer Vision}, pages 250--267. Springer, 2024{\natexlab{b}}.

\bibitem[Li et~al.(2024{\natexlab{c}})Li, Wang, Liu, and Jiang]{binsformer}
Zhenyu Li, Xuyang Wang, Xianming Liu, and Junjun Jiang.
\newblock Binsformer: Revisiting adaptive bins for monocular depth estimation.
\newblock \emph{IEEE Transactions on Image Processing}, 2024{\natexlab{c}}.

\bibitem[Miangoleh et~al.(2021)Miangoleh, Dille, Mai, Paris, and Aksoy]{boost}
S~Mahdi~H Miangoleh, Sebastian Dille, Long Mai, Sylvain Paris, and Yagiz Aksoy.
\newblock Boosting monocular depth estimation models to high-resolution via content-adaptive multi-resolution merging.
\newblock In \emph{Proceedings of the IEEE/CVF Conference on Computer Vision and Pattern Recognition}, pages 9685--9694, 2021.

\bibitem[Moon et~al.(2024)Moon, Bello, Kwon, and Kim]{ground}
Jaeho Moon, Juan Luis~Gonzalez Bello, Byeongjun Kwon, and Munchurl Kim.
\newblock From-ground-to-objects: Coarse-to-fine self-supervised monocular depth estimation of dynamic objects with ground contact prior.
\newblock In \emph{Proceedings of the IEEE/CVF Conference on Computer Vision and Pattern Recognition}, pages 10519--10529, 2024.

\bibitem[Qin et~al.(2022)Qin, Dai, Hu, Fan, Shao, and Van~Gool]{dis}
Xuebin Qin, Hang Dai, Xiaobin Hu, Deng-Ping Fan, Ling Shao, and Luc Van~Gool.
\newblock Highly accurate dichotomous image segmentation.
\newblock In \emph{European Conference on Computer Vision}, pages 38--56. Springer, 2022.

\bibitem[Ramirez et~al.(2023)Ramirez, Costanzino, Tosi, Poggi, Salti, Mattoccia, and Di~Stefano]{booster}
Pierluigi~Zama Ramirez, Alex Costanzino, Fabio Tosi, Matteo Poggi, Samuele Salti, Stefano Mattoccia, and Luigi Di~Stefano.
\newblock Booster: a benchmark for depth from images of specular and transparent surfaces.
\newblock \emph{IEEE Transactions on Pattern Analysis and Machine Intelligence}, 46\penalty0 (1):\penalty0 85--102, 2023.

\bibitem[Ranftl et~al.(2020)Ranftl, Lasinger, Hafner, Schindler, and Koltun]{midas}
Ren{\'e} Ranftl, Katrin Lasinger, David Hafner, Konrad Schindler, and Vladlen Koltun.
\newblock Towards robust monocular depth estimation: Mixing datasets for zero-shot cross-dataset transfer.
\newblock \emph{IEEE transactions on pattern analysis and machine intelligence}, 44\penalty0 (3):\penalty0 1623--1637, 2020.

\bibitem[Ranftl et~al.(2021)Ranftl, Bochkovskiy, and Koltun]{dpt}
Ren{\'e} Ranftl, Alexey Bochkovskiy, and Vladlen Koltun.
\newblock Vision transformers for dense prediction.
\newblock In \emph{Proceedings of the IEEE/CVF international conference on computer vision}, pages 12179--12188, 2021.

\bibitem[Rombach et~al.(2022)Rombach, Blattmann, Lorenz, Esser, and Ommer]{ldm}
Robin Rombach, Andreas Blattmann, Dominik Lorenz, Patrick Esser, and Bj{\"o}rn Ommer.
\newblock High-resolution image synthesis with latent diffusion models.
\newblock In \emph{Proceedings of the IEEE/CVF conference on computer vision and pattern recognition}, pages 10684--10695, 2022.

\bibitem[Saxena et~al.(2008)Saxena, Sun, and Ng]{saxena2008make3d}
Ashutosh Saxena, Min Sun, and Andrew~Y Ng.
\newblock Make3d: Learning 3d scene structure from a single still image.
\newblock \emph{IEEE transactions on pattern analysis and machine intelligence}, 31\penalty0 (5):\penalty0 824--840, 2008.

\bibitem[Scharstein et~al.(2014)Scharstein, Hirschm{\"u}ller, Kitajima, Krathwohl, Ne{\v{s}}i{\'c}, Wang, and Westling]{middle}
Daniel Scharstein, Heiko Hirschm{\"u}ller, York Kitajima, Greg Krathwohl, Nera Ne{\v{s}}i{\'c}, Xi Wang, and Porter Westling.
\newblock High-resolution stereo datasets with subpixel-accurate ground truth.
\newblock In \emph{Pattern Recognition: 36th German Conference, GCPR 2014, M{\"u}nster, Germany, September 2-5, 2014, Proceedings 36}, pages 31--42. Springer, 2014.

\bibitem[Schops et~al.(2017)Schops, Schonberger, Galliani, Sattler, Schindler, Pollefeys, and Geiger]{eth}
Thomas Schops, Johannes~L Schonberger, Silvano Galliani, Torsten Sattler, Konrad Schindler, Marc Pollefeys, and Andreas Geiger.
\newblock A multi-view stereo benchmark with high-resolution images and multi-camera videos.
\newblock In \emph{Proceedings of the IEEE conference on computer vision and pattern recognition}, pages 3260--3269, 2017.

\bibitem[Shao et~al.(2023)Shao, Pei, Chen, Wu, and Li]{shao2023nddepth}
Shuwei Shao, Zhongcai Pei, Weihai Chen, Xingming Wu, and Zhengguo Li.
\newblock Nddepth: Normal-distance assisted monocular depth estimation.
\newblock In \emph{Proceedings of the IEEE/CVF International Conference on Computer Vision}, pages 7931--7940, 2023.

\bibitem[Silberman et~al.(2012)Silberman, Hoiem, Kohli, and Fergus]{nyuv2}
Nathan Silberman, Derek Hoiem, Pushmeet Kohli, and Rob Fergus.
\newblock Indoor segmentation and support inference from rgbd images.
\newblock In \emph{Computer Vision--ECCV 2012: 12th European Conference on Computer Vision, Florence, Italy, October 7-13, 2012, Proceedings, Part V 12}, pages 746--760. Springer, 2012.

\bibitem[Tosi et~al.(2021)Tosi, Liao, Schmitt, and Geiger]{smd}
Fabio Tosi, Yiyi Liao, Carolin Schmitt, and Andreas Geiger.
\newblock Smd-nets: Stereo mixture density networks.
\newblock In \emph{Proceedings of the IEEE/CVF conference on computer vision and pattern recognition}, pages 8942--8952, 2021.

\bibitem[Wang et~al.(2019)Wang, Lucey, Perazzi, and Wang]{wsvd}
Chaoyang Wang, Simon Lucey, Federico Perazzi, and Oliver Wang.
\newblock Web stereo video supervision for depth prediction from dynamic scenes.
\newblock In \emph{2019 International Conference on 3D Vision (3DV)}, pages 348--357. IEEE, 2019.

\bibitem[Wang et~al.(2021)Wang, Zheng, Yan, Deng, Zhao, and Chu]{irs}
Qiang Wang, Shizhen Zheng, Qingsong Yan, Fei Deng, Kaiyong Zhao, and Xiaowen Chu.
\newblock Irs: A large naturalistic indoor robotics stereo dataset to train deep models for disparity and surface normal estimation.
\newblock In \emph{2021 IEEE International Conference on Multimedia and Expo (ICME)}, pages 1--6. IEEE, 2021.

\bibitem[Wang et~al.(2020)Wang, Zhu, Wang, Hu, Qiu, Wang, Hu, Kapoor, and Scherer]{tartanair}
Wenshan Wang, Delong Zhu, Xiangwei Wang, Yaoyu Hu, Yuheng Qiu, Chen Wang, Yafei Hu, Ashish Kapoor, and Sebastian Scherer.
\newblock Tartanair: A dataset to push the limits of visual slam.
\newblock In \emph{2020 IEEE/RSJ International Conference on Intelligent Robots and Systems (IROS)}, pages 4909--4916. IEEE, 2020.

\bibitem[Wang et~al.(2023)Wang, Shi, Li, Huang, Cao, Zhang, Xian, and Lin]{vdw}
Yiran Wang, Min Shi, Jiaqi Li, Zihao Huang, Zhiguo Cao, Jianming Zhang, Ke Xian, and Guosheng Lin.
\newblock Neural video depth stabilizer.
\newblock In \emph{Proceedings of the IEEE/CVF International Conference on Computer Vision}, pages 9466--9476, 2023.

\bibitem[Watson et~al.(2021)Watson, Mac~Aodha, Prisacariu, Brostow, and Firman]{temporal}
Jamie Watson, Oisin Mac~Aodha, Victor Prisacariu, Gabriel Brostow, and Michael Firman.
\newblock The temporal opportunist: Self-supervised multi-frame monocular depth.
\newblock In \emph{Proceedings of the IEEE/CVF conference on computer vision and pattern recognition}, pages 1164--1174, 2021.

\bibitem[Xian et~al.(2018)Xian, Shen, Cao, Lu, Xiao, Li, and Luo]{hr-wsi}
Ke Xian, Chunhua Shen, Zhiguo Cao, Hao Lu, Yang Xiao, Ruibo Li, and Zhenbo Luo.
\newblock Monocular relative depth perception with web stereo data supervision.
\newblock In \emph{Proceedings of the IEEE Conference on Computer Vision and Pattern Recognition}, pages 311--320, 2018.

\bibitem[Xie et~al.(2022)Xie, Xia, Ma, Zhao, Chen, and Li]{uhrsd}
Chenxi Xie, Changqun Xia, Mingcan Ma, Zhirui Zhao, Xiaowu Chen, and Jia Li.
\newblock Pyramid grafting network for one-stage high resolution saliency detection.
\newblock In \emph{Proceedings of the IEEE/CVF conference on computer vision and pattern recognition}, pages 11717--11726, 2022.

\bibitem[Yang et~al.(2024)Yang, Kang, Huang, Xu, Feng, and Zhao]{da}
Lihe Yang, Bingyi Kang, Zilong Huang, Xiaogang Xu, Jiashi Feng, and Hengshuang Zhao.
\newblock Depth anything: Unleashing the power of large-scale unlabeled data.
\newblock In \emph{Proceedings of the IEEE/CVF Conference on Computer Vision and Pattern Recognition}, pages 10371--10381, 2024.

\bibitem[Yang et~al.(2025)Yang, Kang, Huang, Zhao, Xu, Feng, and Zhao]{da2}
Lihe Yang, Bingyi Kang, Zilong Huang, Zhen Zhao, Xiaogang Xu, Jiashi Feng, and Hengshuang Zhao.
\newblock Depth anything v2.
\newblock \emph{Advances in Neural Information Processing Systems}, 37:\penalty0 21875--21911, 2025.

\bibitem[Yin et~al.(2020)Yin, Wang, Shen, Liu, Tian, Xu, Sun, and Renyin]{diverse}
Wei Yin, Xinlong Wang, Chunhua Shen, Yifan Liu, Zhi Tian, Songcen Xu, Changming Sun, and Dou Renyin.
\newblock Diversedepth: Affine-invariant depth prediction using diverse data.
\newblock \emph{arXiv preprint arXiv:2002.00569}, 2020.

\bibitem[Yuan et~al.(2022)Yuan, Gu, Dai, Zhu, and Tan]{newcrf}
Weihao Yuan, Xiaodong Gu, Zuozhuo Dai, Siyu Zhu, and Ping Tan.
\newblock Neural window fully-connected crfs for monocular depth estimation.
\newblock In \emph{Proceedings of the IEEE/CVF conference on computer vision and pattern recognition}, pages 3916--3925, 2022.

\end{thebibliography}
}

\end{document}